%% file: acl_latex.tex
\pdfoutput=1

\documentclass[11pt]{article}

\usepackage{acl}

\usepackage{times}
\usepackage{latexsym}

\usepackage[T1]{fontenc}

\usepackage[utf8]{inputenc}

\usepackage{microtype}

\usepackage{inconsolata}

\usepackage{soul}
\usepackage{amsmath}
\usepackage{booktabs}
\usepackage{bm}
\usepackage{cleveref}
\usepackage{graphicx}
\usepackage{tabularx}
\usepackage{soul}
\usepackage{xcolor}
\usepackage{todonotes}
\usepackage{multirow}
\usepackage{xpatch}
\usepackage{blindtext}
\usepackage{fdsymbol}
\usepackage{microtype}
\usepackage{hyperref}
\usepackage{adjustbox}
\usepackage{textcomp}
\usepackage{xparse}
\usepackage{enumitem}
\usepackage{makecell}
\usepackage{subcaption}
\usepackage{colortbl}

\usepackage{pifont}

\usepackage{xparse}
\usepackage{stfloats}

\crefformat{section}{\S#2#1#3} %
\crefformat{subsection}{\S#2#1#3}
\crefformat{subsubsection}{\S#2#1#3}

\input{contents/commands}

\title{Embrace Divergence for Richer Insights: \\A Multi-document Summarization Benchmark and a Case Study on Summarizing Diverse Information from News Articles}

\author{Kung-Hsiang Huang$^{1}$\thanks{~~Work done while interning at Salesforce AI Research.} ~~~Philippe Laban$^{2}$ ~~~ Alexander R. Fabbri$^{2}$ \\ {\bfseries ~~~ Prafulla Kumar Choubey$^{2}$ ~~~ Shafiq Joty$^{2}$  ~~~ Caiming Xiong$^{2}$ ~~~ Chien-Sheng Wu$^{2}$}\\
$^{1}$University of Illinois Urbana-Champaign ~~~ $^{2}$Salesforce AI Research\\
$^{1}$\texttt{khhuang3@illinois.edu} \\
$^{2}$\texttt{\{plaban, afabbri, pchoubey, sjoty, cxiong, wu.jason\}@salesforce.com}}

\PassOptionsToPackage{table}{xcolor}

\begin{document}
\maketitle
\input{contents/00_abstract}

\input{contents/01_introduction}

\input{contents/03_task}

\input{contents/04_data}

\input{contents/05_evaluation}

\input{contents/02_related_work}

\input{contents/06_conclusion}
\input{contents/07_ethics}
\input{contents/08_limitation}
\bibliography{anthology,custom}
\bibliographystyle{acl_natbib}

\input{contents/appendix}

\end{document}

%% file: contents/commands.tex
\newcommand{\samsum}[1]{\textsc{SAMSum}}
\newcommand{\dialogsum}[1]{\textsc{DialogSum}}
\newcommand{\mixandmatch}[1]{\textsc{MixAndMatch}}
\newcommand{\confit}[1]{\textsc{ConFiT}}
\newcommand{\ctrldiasumm}[1]{\textsc{CtrlDiaSumm}}
\newcommand{\cods}[1]{\textsc{CODS}}
\newcommand{\modelshort}[1]{\textsc{ZeroFEC}}
\newcommand{\modelshortda}[1]{\textsc{ZeroFEC-DA}}
\newcommand{\datashort}[1]{\textsc{DiverseSumm}}
\newcommand{\tasklong}[1]{Multi-document Diversity Summarization}
\newcommand{\taskshort}[1]{MDDS}
\newcommand{\gptturbo}[1]{GPT-3.5-Turbo}
\newcommand{\gptturbolong}[1]{GPT-3.5-Turbo-16K}
\newcommand{\gptfour}[1]{GPT-4}
\newcommand{\vicuna}[1]{Vicuna-7B}
\newcommand{\longchat}[1]{LongChat-7B-16K}
\newcommand{\CC}[1]{\cellcolor{lightblue!#1}}

\definecolor{c2}{RGB}{218,0,0}

\definecolor{lightblue}{RGB}{212, 235, 255}
\definecolor{lightorange}{RGB}{255, 204, 168}
\definecolor{lightyellow}{RGB}{255, 255, 168}
\definecolor{lightred}{RGB}{255, 168, 168}
\definecolor{darkred}{RGB}{196, 30, 58}
\definecolor{lightgreen}{rgb}{0.82, 0.94, 0.75}
\definecolor{lightpurple}{RGB}{225, 213, 231}
\definecolor{darkgreen}{rgb}{0.56, 0.63, 0.51}
\definecolor{normalgreen}{rgb}{0.66, 0.9, 0.5}
\definecolor{lightgray}{rgb}{0.7, 0.7, 0.7}
\definecolor{gold}{rgb}{0.83, 0.69, 0.22}
\sethlcolor{lightblue}
\newcolumntype{Y}{>{\centering\arraybackslash}X}
\newcommand\hlc[2]{\sethlcolor{#1} \hl{#2}}
\definecolor{gold}{rgb}{0.83, 0.69, 0.22}

\newcommand{\cmark}{\ding{51}}%
\newcommand{\xmark}{\ding{55}}%

%% file: contents/00_abstract.tex
\begin{abstract}
    Previous research in multi-document news summarization has typically concentrated on collating information that all sources agree upon. However, the summarization of diverse information dispersed across multiple articles about an event remains underexplored. In this paper, we propose a new task of summarizing diverse information encountered in multiple news articles encompassing the same event. To facilitate this task, we present a data collection schema for identifying diverse information and curated a dataset named \datashort~. The dataset includes 245 news stories, with each story comprising 10 news articles and paired with a human-validated reference. Next, to enable consistent automatic evaluation, we conduct a comprehensive analysis to pinpoint the position and verbosity biases when utilizing Large Language Model (LLM)-based metrics for evaluating the coverage and faithfulness of summaries. Through correlation analyses, we outline the best practices for effectively using automatic LLM-based metrics on the \datashort~ dataset. Finally, we study how LLMs summarize multiple news articles by analyzing which type of diverse information LLMs are capable of identifying. Our analyses suggest that despite the extraordinary capabilities of LLMs in single-document summarization,  the proposed task remains a complex challenge for them mainly due to their limited coverage, with \gptfour~ only able to cover under 40\% of the diverse information on average.\footnote{The code and data have been made publicly available:
\url{https://github.com/salesforce/DiverseSumm}.}

\end{abstract}

%% file: contents/01_introduction.tex
\section{Introduction}

In the realm of news reporting, each event is often chronicled by multiple sources, providing a rich tapestry of perspectives and insights. The sheer volume of articles available via news aggregators, as noted by \citet{laban2023designing}, can overwhelm readers, leading to fatigue \cite{lee2015rise}. This has fueled the demand for more digestible multi-source summaries. However, as highlighted by existing multi-document summarization studies \cite{over2004introduction, owczarzak2011overview, fabbri-etal-2019-multi}, these often only reflect consensus information and neglect the breadth of differing viewpoints. To address this, we propose the \textbf{\tasklong~ (\taskshort~)} task, aimed at faithfully illuminating the diverse information presented in multiple sources. \looseness=-1

\begin{figure*}[bt]
    \centering
    \includegraphics[width=0.97\linewidth]{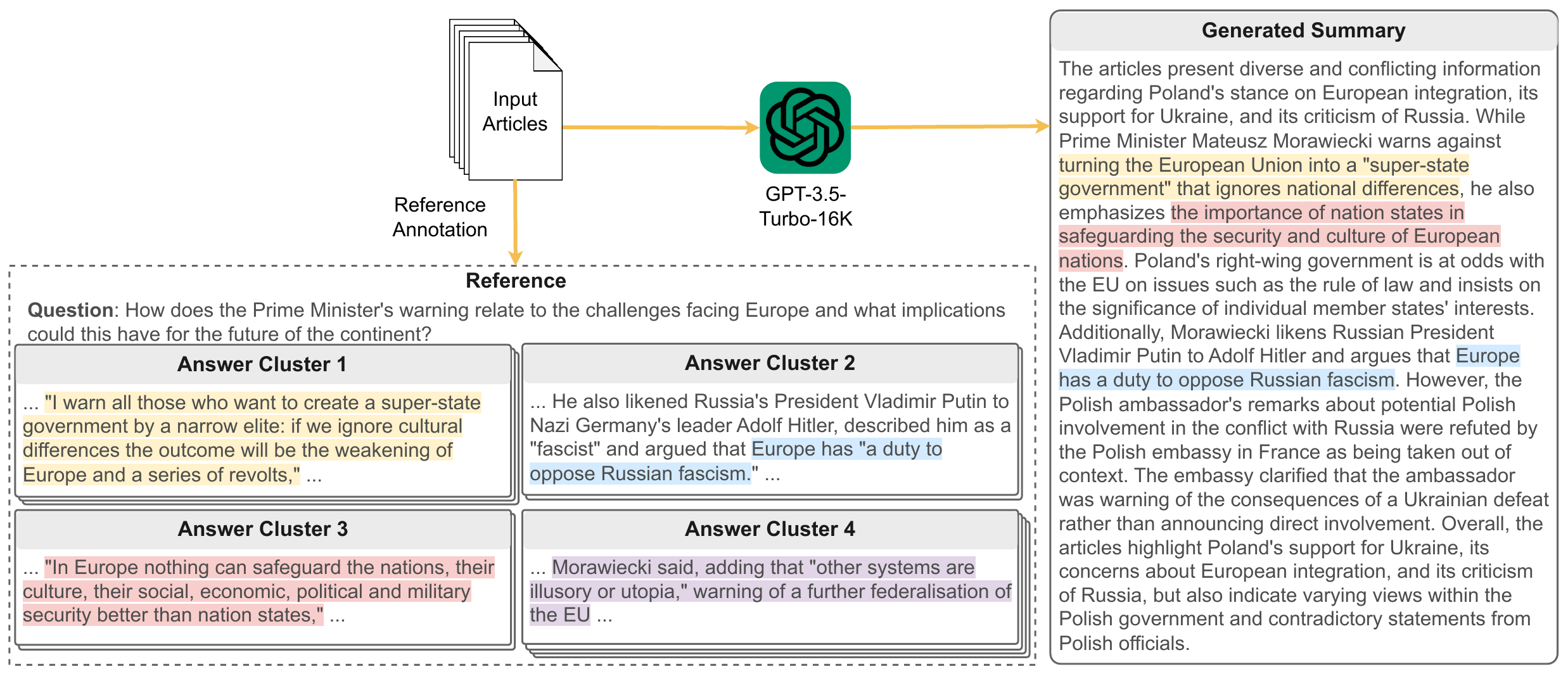}
    \vspace{-2mm}
    \caption{An example from our \datashort~ dataset and a summary generated by \gptturbolong~. To depict the process succinctly, only 4 news answer clusters from the reference are displayed. In this instance, the reference contains a single question with various answers extracted from each news article. In general, a news event may contain multiple reference questions, each of which can correspond to multiple answer clusters. The summary produced by \gptturbolong~ encompasses 3 of the answer clusters shown, but does not cover \hlc{lightpurple}{Answer Cluster 4}.\looseness=-1} %
    \label{fig:toy_example}
    \vspace{-5mm}
\end{figure*}

Following \citet{laban-etal-2022-discord}, we formalize diverse information as \textit{questions and answers where numerous sources can answer the same question, and the corresponding answers extracted from different news articles exhibit a variety of opinions or perspectives}. For robust and objective evaluation, we opted for a QA representation for references, aligning with the granularity and reliability advantages emphasized in prior work on summarization evaluation \cite{krishna-etal-2023-longeval, liu-etal-2023-revisiting, arumae-liu-2019-guiding}. An example of diverse information is shown in \Cref{fig:toy_example}.

Using this formulation, we propose a reference annotation methodology to identify and gather diverse information dispersed across multiple articles about the same story. Our approach is a pipeline based on \gptturbo~ \cite{openai2023chatgpt}, which generates questions concerning the story likely to pull varied responses from different sources. The subsequent answers extracted from each news article are then clustered into groups. We employ a post-processing step that removes invalid questions and answers. Finally, all questions and answers are validated by human annotators. The resulting dataset contains 245 news story clusters, where each story contains 10 news articles and an average of 2.49 questions, with each question associated with 3.41 answer clusters on average. This dataset is named \textbf{\datashort~}.

We conduct a series of experiments to understand the relevancy and challenges of our task in the era of LLMs and how future work should evaluate models on our task. Our fine-grained human evaluation results identify that even the most advanced LLM, \gptfour~, only covers about 37\% of diverse information with optimally designed prompts (see \Cref{apx:llm_prompts_summ}). This highlights the significant challenge of effectively incorporating diverse information from multiple sources and the efficacy of our dataset as a rigorous LLM benchmark. Furthermore, we assess \gptfour~ as an evaluator, given the impracticality of extensive human evaluations and its high correlation with human ratings \cite{liu2023gpteval}. Based on the correlation and bias analysis of \gptfour~ evaluations, we provide recommendations for its application in assessing coverage and faithfulness of LLMs on our task. Our key findings are outlined in \Cref{tab:rq_summary}.

Our contributions are: (1) We introduce the \tasklong~ task that challenges models to identify diverse information across news articles and propose a reference annotation scheme to construct the \datashort~ dataset. (2) We conduct extensive human evaluations to understand LLMs' ability to tackle our task and demonstrate that even \gptfour~ struggle to achieve high coverage. (3) We conduct bias and correlation analysis on different \gptfour~-based evaluation protocols to provide recommendations on using \gptfour~-based metrics on our task. These guidelines are used to assess the coverage bias in various LLMs to understand how they summarize diverse information, highlighting the remaining challenges.

\label{sec:intro}

%% file: contents/03_task.tex
\input{tables/rq_summary}

\section{Task}

The \taskshort~ task revolves around a cluster of $K$ news articles all centered around the same news event. To maintain a balance between task feasibility and challenge, we have opted to set $K$ at a value of 10. The primary aim of our task is to generate a natural-language summary that effectively captures the diverse information presented within this cluster of news articles. To facilitate this process, our data collection pipeline, as elaborated in \Cref{sec:data}, produces references for each news cluster. These references take the form of question-answers (QAs), and their validity is established through human validation. The QAs must satisfy two properties: (1) the valid question must be answered by a sufficient number of sources, and (2) the answers associated with a valid question must present diverse opinions or perspectives.

In this work, the assessment of the generated summaries centers on two key facets: faithfulness and coverage. The faithfulness aspect evaluates the extent to which the summary aligns with the factual content present in the source articles. On the other hand, the coverage aspect gauges the inclusivity of information by considering how many answers within the reference are effectively addressed in the summary. We set our primary focus on these two aspects instead of other qualities, such as compression ratio and coherence, because recent work has shown that faithfulness and coverage are two major summarization challenges faced by models based on pre-trained transformers \cite{cao-wang-2021-cliff, tang-etal-2022-confit, huang-etal-2023-swing, qiu-etal-2024-amrfact}. \looseness=-1

%% file: tables/rq_summary.tex
\begin{table*}[t]
    \small
    \centering
    \begin{adjustbox}{max width=0.98\textwidth}
    {
    \begin{tabular}{p{0.98\linewidth}}
        \toprule
        
        \textbf{RQ1:  How proficient are LLMs in summarizing diverse information from multiple news articles about an event?}\\
           
        \midrule
        
        - While LLMs can generate faithful summaries, they often lack adequate coverage. \\
        - Given the challenge of multi-document diverse summarization, our dataset serves as a rigorous benchmark for LLMs.\\
        \midrule
        \textbf{RQ2: What are the pitfalls and best practices when leveraging \gptfour~ as the evaluation metric for our task?}\\
        \midrule
        - As a pairwise evaluator, \gptfour~ shows a bias for the second summary.\\
        - Used as a single-answer grader, \gptfour~ is prone to verbosity bias and prefers shorter summaries.\\
        - Likert-scale grading balances budget with correlation to human judgment for faithfulness evaluation. \\
        - Both granular evaluation methods correlate well with human judgment for coverage.\\

        \midrule
        \textbf{RQ3: Do LLMs exhibit coverage bias when performing \taskshort~?}\\
        \midrule
        - LLMs usually focus on summarizing the initial and final input articles, often overlooking the middle ones.\\
        - LLMs struggle to comprehensively address "How" and "What" type questions. \\
        - Long-context LLMs excel at covering frequent answers, while standard LLMs are proficient at summarizing infrequent ones.\\
        - Increasing model size improves LLMs' coverage of diverse information. \\

        \bottomrule
    \end{tabular}
    }
    \end{adjustbox}
    \vspace{-2mm}
    \caption{Summary of research questions and key findings of our study.} 
    \label{tab:rq_summary}
    \vspace{-5mm}
\end{table*}

%% file: contents/04_data.tex
\section{Data Collection}
\label{sec:data}
This section details the \datashort~ data collection pipeline, delineating its automated diverse information discovery from articles and the human validation stage that ensures data integrity.
\subsection{Automatic Data Curation}
\label{subsec:automatic_data_collection}
Our data collection framework surfaces diverse information across news articles by asking questions about a news story, extracting answers from each news article, clustering the answers based on semantics, and filtering invalid questions and answers that are invalid. Our method extends the Discord Questions data generation pipeline \cite{laban-etal-2022-discord} with four major modifications aimed at improving data quality: 

(1) We perform question generation in a two-stage fashion, which increases the number of questions that result in diverse answers extracted from different articles. (2) Our question-answering component extracts answers from the context of the entire article, instead of extracting from each paragraph independently, significantly improving the recall of answers. (3) We perform a post-processing step to remove answers that do not make sense and QA-pairs that do not form diverse information. (4) Our method is based on \gptturbo~ \footnote{We used the \texttt{gpt-3.5-turbo-0613} variant.}, allowing for collection of higher-quality data. \looseness=-1

\paragraph{Data Source}
We create \datashort~ by gathering news stories and corresponding events from Google News, a news aggregator that collects news articles from various sources for a given news story. Each news story in Google News corresponds to around 40 news articles. We picked 400 news stories on the recent section of Google News. Most articles were published during March 2023, hence beyond the knowledge cut-off date of \gptturbo~, which is September 2021. %

\paragraph{Question Generation}
Upon collecting news stories, our next step is to ask questions about each news story that satisfy two properties: (1) \textit{Availability of response}: this property ensures that any question deemed valid for the task should be one that many source articles can answer, hence indicating its centrality to the news event being reported. It is about the presence of answers across the corpus rather than their content.
(2) \textit{Diversity of answers}: this property focuses on the content of the responses rather than their presence. It stipulates that the answers to a valid question should exhibit a range of perspectives or opinions when extracted from different sources/articles. This is the heart of our approach to capturing the diversity of viewpoints represented in news articles.

We validate a query if at least 30\% of the sources answer it and it results in assorted responses. To assess the efficiency of various methods of Question Generation (QG), we manually reviewed 10 news stories. We extend the Discord Question framework \cite{laban-etal-2022-discord} by replacing their QG component with \gptturbo~ for its better performance over smaller models. For each news narrative, we heuristically select a medium-length article to prompt \gptturbo~, generating 20 questions each, after which answers are extracted from all sources using the QA method outlined subsequently. The analysis reveals that of the 200 questions generated via this method, only 42 questions sufficiently cover all source articles, with a mere 10 questions satisfy the two requirements mentioned above, indicating the single-article input's limited recall. \looseness=-1

To enhance question coverage, we incorporate multiple representative articles into \gptturbo~. We hypothesize that the answer clusters identified by a RoBERTa-based QA pipeline \cite{laban-etal-2022-discord} provide a decent degree of diversity. 
Consequently, we identified representative articles through a heuristic method: a question corresponding to the median number of answer clusters was chosen. Within the associated articles, we opted for a medium-length article. This process produces a set of representative articles for the chosen questions corresponding to a news story. Prompting \gptturbo~ with these articles yielded 20 questions. \looseness=-1%

On a manual assessment of the aforementioned 10 news stories, this novel approach increased the number of questions linked with sufficient answers and valid questions, to 85 (+102.4\%) and 19 (+90.0\%), respectively. 
This indicates the proposed QG strategy's efficacy, significantly increasing the generation of valid questions compared to the prior method \cite{laban-etal-2022-discord}, and justifies our hypothesis mentioned in the previous paragraph. %

\paragraph{Question Answering}

Similar to QG, we create an evaluation set for assessing the performance of question answering (QA) on our collected data, which contains two news stories, each paired with six human-generated valid questions. We compared various QA models, including a RoBERTa-based model \cite{liu2019roberta} and two \gptturbo~ variants. One \gptturbo~ variant processes paragraphs independently, akin to RoBERTa, while its article-level counterpart extracts answers from the entire news article. Upon inspecting the outputs, we found that RoBERTa demonstrated higher precision, but the article-level \gptturbo~ variant excelled in recall (64.6\%) against RoBERTa's (43.8\%). Given the ease of filtering excessive answers compared to recovering missed answers, we opted for the article-level \gptturbo~ for all subsequent experiments.\looseness=-1 %

\paragraph{Answer Consolidation}
For answer consolidation, we conduct a similar small-scale analysis to understand the performance of different answer clustering methods. We do not find significant advantages of the method based on \gptturbo~ compared to prior approaches; hence, we use the RoBERTa-based method \cite{laban-etal-2022-discord} as our answer consolidation model. \looseness=-1

\paragraph{Post-processing}

To ensure task feasibility, we downsize the articles by selecting articles that have higher coverage of answers such that each news story is now associated with at most 10 articles. To expedite the process of human validation illustrated in \Cref{subsec:human_validation}, we utilized \gptturbo~ to filter non-sensical answers and non-diverse QA-pairs. Questions that are no longer associated with adequate answers due to the filtering are removed. Similarly, news stories that do not have any valid questions because of the filtering will be removed as well. The LLM prompts used in this subsection can be found in \Cref{apx:llm_prompts_data}.

\subsection{Human Validation}
\label{subsec:human_validation}
\begin{figure}[t]
    \centering
    \begin{subfigure}{0.45\textwidth}
    \centering
    \includegraphics[width=.95\linewidth]{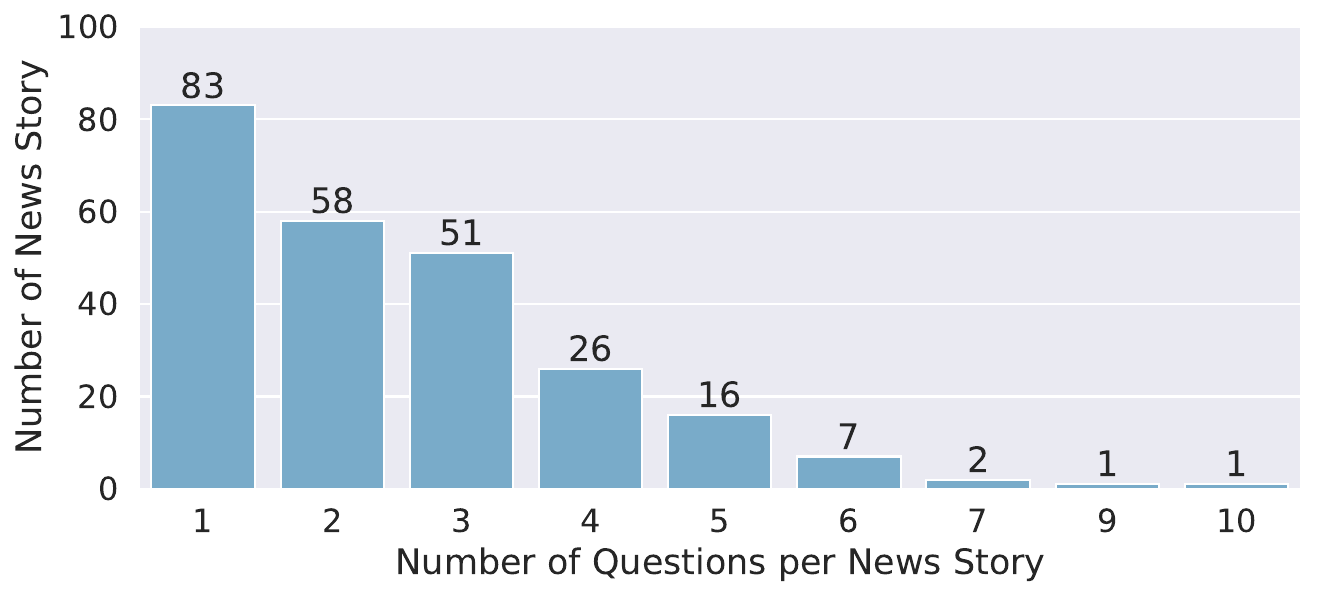}
    
    \end{subfigure}
    ~
    \begin{subfigure}[b]{0.45\textwidth}
    \centering
    \includegraphics[width=.95\linewidth]{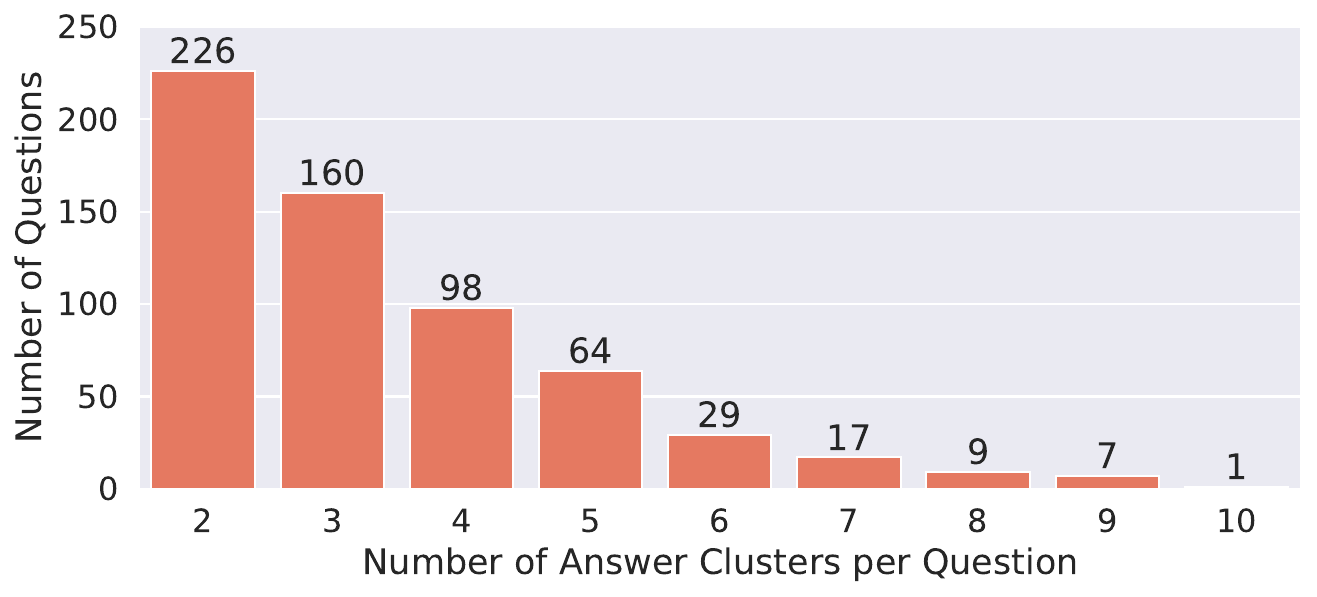}
    \end{subfigure}
        
    \vspace{-2mm}
    \caption{Dataset statistics regarding the number of questions and answer clusters.\looseness=-1}%
    \label{fig:dataset_stats}
    \vspace{-5mm}
\end{figure}

To address any invalid QA-pairs that slipped past our post-processing procedure and enhance data quality, we recruited human annotators to validate the post-processed QAs. They are tasked to verify whether an answer addresses the corresponding question and ensure at least one article contains such an answer. More about this process is detailed in \Cref{apx:human_annotation_qas}. The resulting \datashort~ dataset contains 245 news stories, each containing 10 articles. The distribution of the number of questions per news story and the number of answer clusters per question are shown in \Cref{fig:dataset_stats}. The distribution of question types and the topic of these news stories are shown in \Cref{apx:topic_distribution}. %

%% file: contents/05_evaluation.tex
\input{tables/llm_performance}
\section{Analysis}

We address the research questions from \Cref{sec:intro}, first evaluating how well diverse information from multiple sources is summarized by LLMs (\Cref{subsec:rq1}), then examining LLM behavior during this summarization (\Cref{subsec:req3}) using the most reliable LLM-based evaluation protocols we found (\Cref{subsec:req2}). 

\subsection{RQ 1: How proficient are LLMs in summarizing diverse information from multiple news articles?}
\label{subsec:rq1}

To understand LLMs' performance on \taskshort~, we conduct human evaluation on summaries produced by four representative LLMs, \gptfour~ \cite{openai2023gpt4}, \gptturbolong~ \cite{openai2023gpt4}, \vicuna~ \cite{vicuna2023}, \longchat~ \cite{longchat2023}.\footnote{We use \texttt{gpt-4-0613}, \texttt{gpt-3.5-turbo-16k-0613}, \texttt{vicuna-7b-v1.3} and \texttt{longchat-7b-16k}.} \textit{Long-context} LLMs, \gptturbolong~ and \longchat~, handle texts up to 16K tokens and can perform direct summarization by taking all articles as input. \textit{Standard LLMs}, \gptfour~ and \vicuna~, are limited to 8K and 2K tokens, respectively; hence, we split summarization into two stages: selecting the most salient $N$ sentences from each article and summarizing these sentences.\footnote{We chose $N=5$.}  To elicit a high-coverage summary of diverse information, we manually optimize the prompts. Details of the prompts used for summarization in our experiments can be found in \Cref{apx:llm_prompts_summ}. Following \citet{krishna-etal-2023-longeval}, we conduct evaluations at a finer granularity. Faithfulness is judged per sentence, whereas coverage is determined by how many reference QA pairs are covered by each summary. The resultant scores for each LLM were averaged from evaluations per summary sentence and reference QA pair, respectively. Evaluation details, such as worker qualification and user interface, are in \Cref{apx:human_eval}.

The human evaluation results are presented in \Cref{tab:llm_performance_human}. We observe that all four LLMs in general achieve high faithfulness but insufficient coverage of diverse information. This suggests that the proposed task is challenging even for state-of-the-art LLMs, and highlights that \datashort~ serves as a challenging test bed for LLMs. %

\input{tables/position_bias_analysis}

\input{tables/verbosity_bias_analysis}

\subsection{RQ 2: What are the pitfalls and best practices when leveraging GPT-4 as the evaluation metric for our task?}
\label{subsec:req2}
To facilitate the analysis and discussion of our next research question, we rely on LLM-based evaluation metrics to conduct various analyses, given their superior correlation with human judgments \cite{liu2023gpteval} and the high cost of human annotation. For this research question, we aim to provide the best practices when using \gptfour~ as the evaluator for the \taskshort~ task by conducting bias and correlation analyses. %

We focus on two major biases: position bias (i.e., whether the LLM evaluator favors certain positions over others) and verbosity bias (i.e. whether the LLM evaluator prefers shorter or longer texts). %
For all the experiments conducted in this analysis, we investigated summaries produced by \gptfour~, \gptturbo~, \vicuna~, and \longchat~. The details of our prompts for the below experiments can be found in \Cref{apx:llm_prompts_eval}.

\paragraph{Position Bias} Position bias is most relevant to the pairwise comparison protocol. While previous work has shown that \gptfour~ does exhibit position bias when used to assess text quality in conversational-focused tasks \cite{wang2023large, zheng2023judging}, none of the prior studies have investigated whether such bias is also observed when evaluating faithfulness or coverage. To analyze position bias, we task \gptfour~ with assessing a pair of summaries generated by two LLMs on which one is better, and then swap the positions of these two summaries and query \gptfour~ again. We compute the percentage of times \gptfour~ prefers the first or second summaries.\looseness=-1

When \gptfour~ compared pairs of LLM-generated summaries to evaluate faithfulness and coverage, a strong position bias surfaced, favoring the second entry (\Cref{tab:position_bias_analysis}). Position bias was particularly pronounced when assessing similar-quality summaries (see \Cref{fig:position_bias_pairwise_coverage}a). Hence, we deduce that \textbf{\gptfour~ is unreliable when utilized as a pairwise evaluator in the \taskshort~ task with respect to faithfulness and coverage}. Interestingly, this outcome contradicts \citet{zheng2023judging}, implying that \textbf{the position of bias for LLM-based evaluators could vary across different tasks}. A breakdown of the position bias analysis can be found in \Cref{apx:llm_bias_analysis}.

\input{tables/protocol_correlation}

\paragraph{Verbosity Bias}

To assess the verbosity bias of \gptfour~ as an evaluator, we create extended summaries that maintain the semantic meaning. We achieve this by duplicating the original summaries, following \citet{zheng2023judging}. Ideally, a fair evaluator should provide identical faithfulness and coverage scores for both the original and extended summaries. %
We employed two experimental designs: pairwise comparison and single-answer grading on a Likert scale of 5.  %

The results of our verbosity bias analysis can be found in \Cref{tab:verbosity_bias_analysis}. We see that when \textbf{using the single-answer grading protocol, \gptfour~ has a strong preference over shorter summaries, whether it is assessing faithfulness or coverage}. This conclusion was unexpected, particularly as we anticipated \gptfour~ to favor longer summaries when determining coverage. Additionally, we noted that \textbf{verbosity bias is significantly lessened when using the pairwise comparison protocol}, which also comes with a much higher computational cost.\looseness=-1

\paragraph{Correlation Analysis}

Upon examining the biases, we explore LLM-based evaluation protocols for their alignment with human judgments, varying reference granularity and rating models, including the use of \gptturbolong~ for efficiency in faithfulness assessment.
For the pairwise comparison, since we had already established the prevalence of its significant position bias, we conducted the comparison both ways by swapping the summaries and then aggregating the results. As shown in \Cref{tab:protocol_correlation}, the both-way pairwise comparison protocol highly correlate with human judgment, mitigating verbosity and position biases, but was computationally demanding. In contrast, single-answer document-summary grading was efficient and fairly accurate. Notably, some \gptturbolong~ protocols negatively correlate with human assessment, indicating that \textbf{even though state-of-the-art long-context LLMs have a wide context window, their capacity to reason through extensive text effectively is occasionally unsatisfactory}.

In terms of coverage, we observed that both coarse-grained (QA-pairs) and fine-grained (single QA) evaluation protocols can establish a reasonably high correlation with human judgments provided we use appropriate rating methods (i.e., Likert scale for the former and binary rating for the latter). Either protocol proves suitable, contingent upon the level of granularity required for analysis.

\paragraph{Evaluation Recommendations}
For faithfulness evaluation, if budget is not a concern, it is recommended to use both-way pairwise comparisons given its high correlation with human judgments and least bias (The average cost for this evaluation protocol on our dataset is around \$200 for each pair of models.). Otherwise, Likert scale single-answer grading with \gptfour~ is the optimal alternative. For coverage evaluation, Likert scale single-answer grading has the highest correlation with human judgments.

\begin{figure}[t]
    \centering
    \begin{subfigure}{0.45\textwidth}

    \includegraphics[width=.95\linewidth]{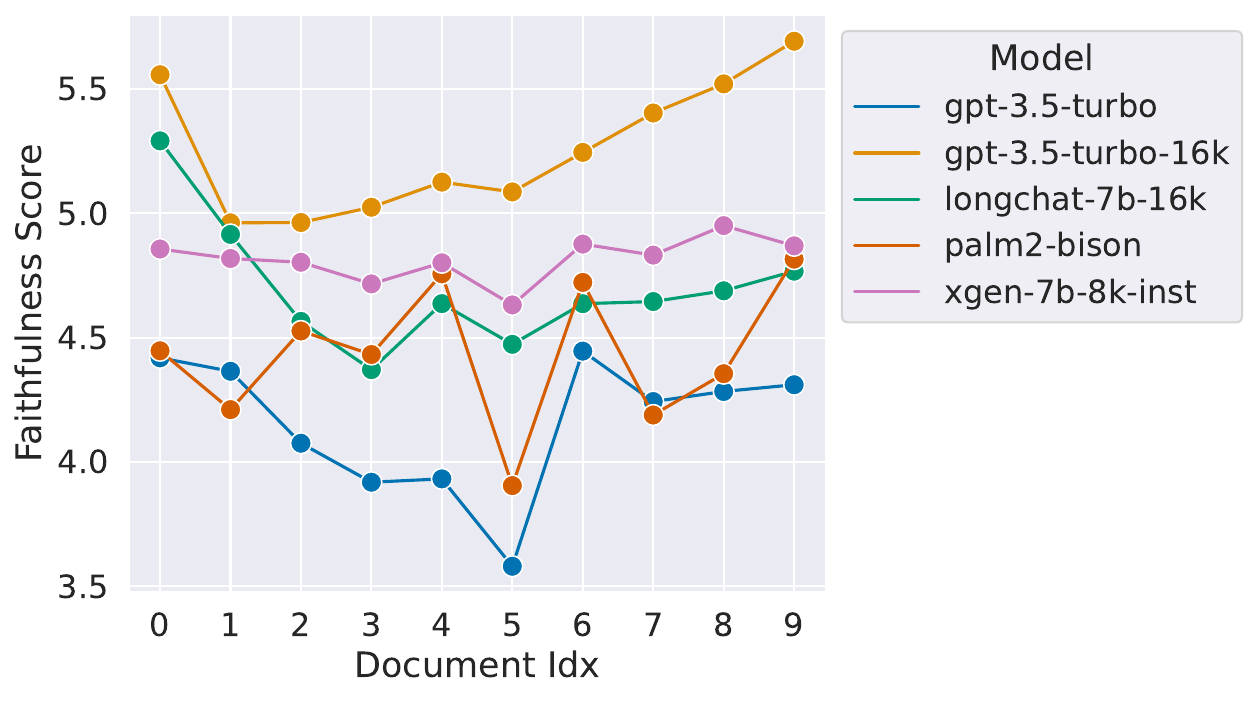}
    
    \end{subfigure}
    ~
    \begin{subfigure}[b]{0.45\textwidth}
    \includegraphics[width=.95\linewidth]{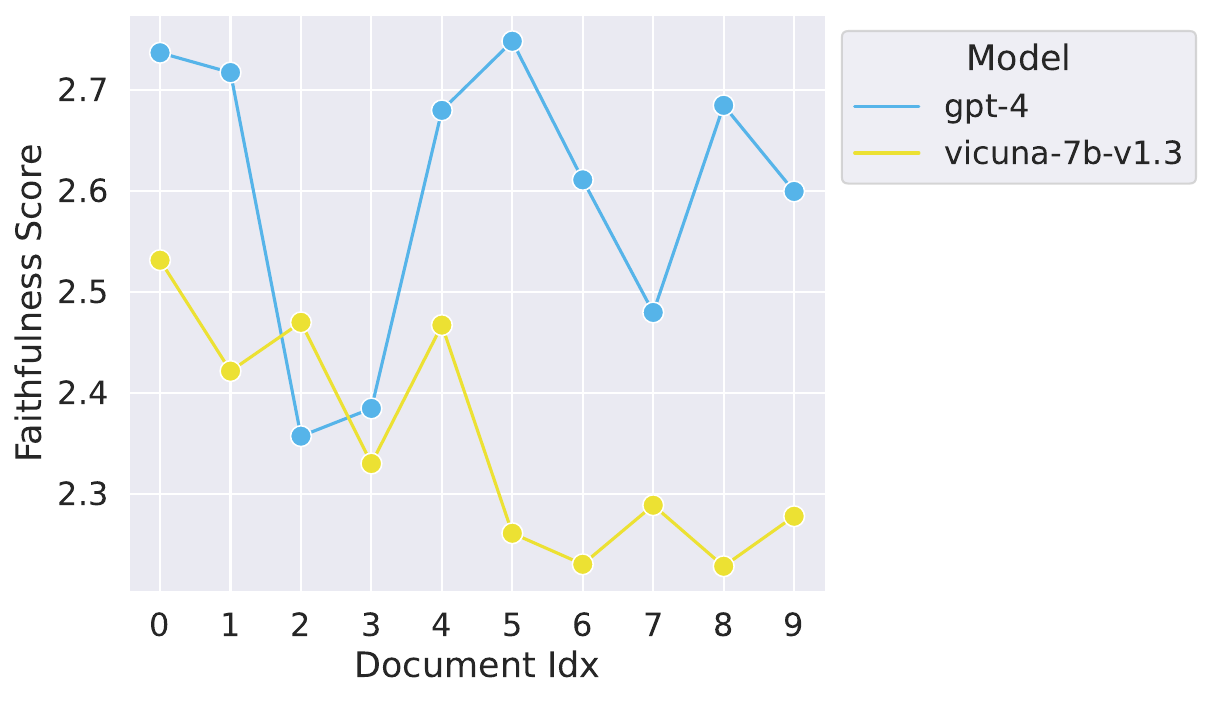}
    \end{subfigure}
        
    \vspace{-2mm}
    \caption{Faithfulness scores w.r.t. the index of the news article in the input prompt for LLMs. We see that LLMs with higher faithfulness (top), regardless of the way it summarize the article, tend to summarize from the starting or ending articles, while such a pattern is not observed for LLMs of low faithfulness (bottom).\looseness=-1}
    \label{fig:rq3_position}
    \vspace{-5mm}
\end{figure}

\subsection{RQ 3: Do LLM exhibit coverage bias when performing \taskshort~? }%
\label{subsec:req3}
With the insights drawn from our analysis of the previous research questions, we are able to effectively conduct experiments to answer what type of information LLMs tend to summarize. We break down this research question into three sub-questions, with each focus on different aspects: focusing on article position, question type, and answer frequency. 
Since the evaluation is automatically conducted using \gptfour~, we additionally consider the following LLMs for analysis: \gptturbo~, XGen-7B-8K-Inst \cite{xgen}, and Palm2-Bison \cite{palm2bison}. The results are discussed in the following paragraphs. %

\paragraph{Do LLMs tend to summarize articles at particular positions?}

The faithfulness score can serve as a measure to gauge how much content in an article's summary is drawn from each input news article. Higher faithfulness indicates greater information extraction from corresponding articles. %
We compute the faithfulness score between the generated summaries and each corresponding article using \gptfour~ based on the article-summary Likert-scale single-answer grading protocol. In \Cref{fig:rq3_position}, a prominent U-shape pattern for faithful LLMs (top) suggests that \textbf{faithful LLMs tend to summarize content from the first and last articles, while giving less attention to the middle articles}, aligning with findings from \citet{liu2023lost} on QA tasks. However, lower-faithfulness LLMs (bottom) show no clear pattern.\footnote{\gptfour~'s lower faithfulness scores arise from their summaries containing article indexes, which are not presented to the evaluators during the evaluation process.}%

\begin{figure}[bt]
    \centering
    \includegraphics[width=0.95\linewidth]{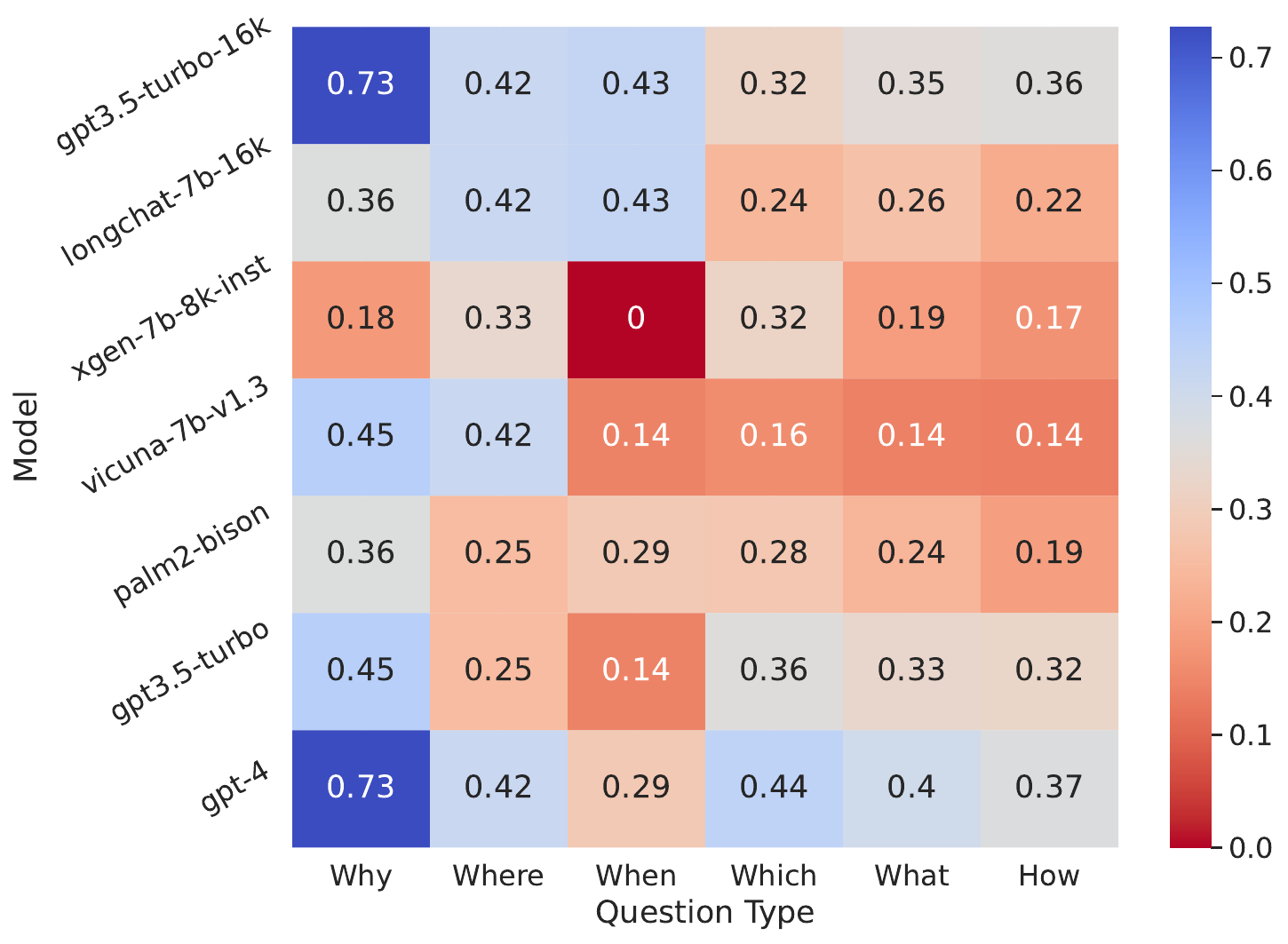}
    \vspace{-2mm}
    \caption{Average coverage scores with regard to different question types for different LLMs. Blue indicates a higher coverage, while red represents a lower coverage.\looseness=-1} 
    \label{fig:rq3_question}
    \vspace{-5mm}
\end{figure}

\paragraph{What diverse information do LLMs best identify and summarize?}

To understand categories of diverse information that LLMs are more inclined to summarize, we analyzed coverage by question type, with each binary coverage score mapping a summary to reference answers  using \gptfour~ with the QA-summary binary single-answer grading protocol. Then, we aggregate these answers based on the respective question types and calculate the averages, as depicted in \Cref{fig:rq3_question}. Results show that questions starting with ``Why'' and ``Where'' tend to have better coverage, likely due to the direct presence of related answers in the source articles. Conversely, \textbf{LLMs encounter challenges in adequately covering answers for ``How'' and ``What'' type questions.} These question types delve into implications and require the model to establish connections between events, making them more intricate to address. Two examples of different types of questions are demonstrated in \Cref{tab:qualitative_examples_qa}. %

\input{tables/qualitative_examples_qa}

\begin{figure}[t]
    \centering
    \includegraphics[width=0.95\linewidth]{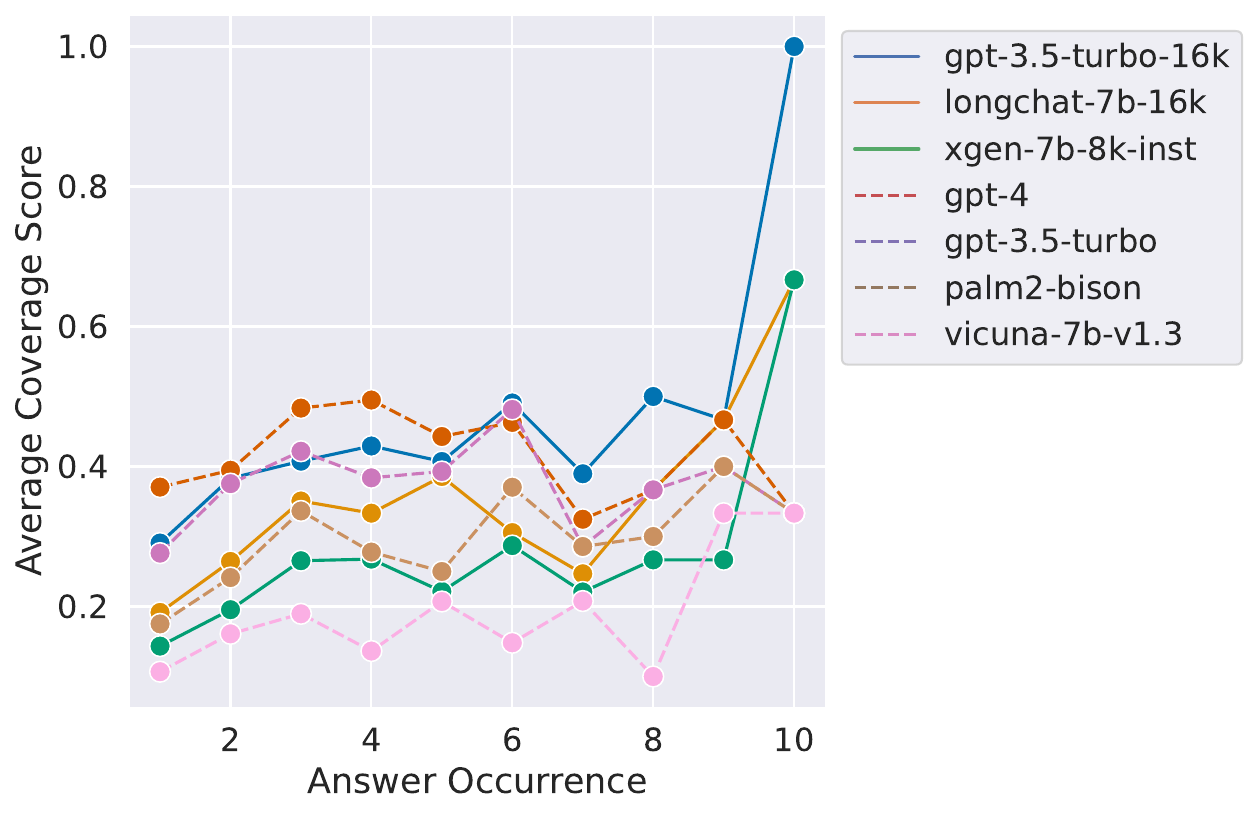}
    \vspace{-2mm}
    \caption{Average coverage scores with regard to answer frequency for different LLMs. Solid lines denote long-context LLMs, while dotted lines indicate standard LLMs. Answer occurrence represents the number of articles containing a given answer. For example, an answer occurrence of 10 means that all 10 input articles contain such an answer. \looseness=-1}  %
    \label{fig:rq3_frequency}
    \vspace{-5mm}
\end{figure}
\paragraph{Do LLMs have a tendency to summarize frequent information?}

We are intrigued by how the frequency of a piece of information influences the behavior of LLMs when summarizing multiple articles. Our data collection approach has facilitated this analysis, as answers extracted from each article have been systematically grouped. This enables us to easily determine the occurrence of a specific answer by calculating the number of articles containing that particular answer within its cluster. We compute the average coverage scores by aggregating answers based on their frequency of occurrence. The results, illustrated in \Cref{fig:rq3_frequency}, reveal a notable trend: frequent answers (i.e., those found in a higher number of articles) tend to be covered more. Additionally, we found that \textbf{long-context LLMs exhibit greater proficiency in covering frequent answers, while standard LLMs appear to excel at summarizing infrequent answers}. This distinction is evident in the comparison between the performance of \gptfour~ and \gptturbolong~. \looseness=-1

\input{tables/size_coverage}

\paragraph{Does the size of LLMs correlate with their coverage of diverse information?}
To run this analysis, we need to ensure that factors other than the size of the model do not influence the results. Hence, we conduct experiments with LLMs in the same family. These include a family of \textit{standard} LLMs, Llama-2 \cite{touvron2023llama}, with a maximum token length of 4K, as well as a family of \textit{long-context} LLMs, Vicuna-v1.5-16K%
, which can handle up to 16K tokens. To measure the coverage scores, we utilized the evaluation protocol that shows the highest correlation with human judgment, as shown in \Cref{tab:protocol_correlation}. This consisted of a single-answer Likert-scale grading scheme, using question-and-answer pairs as the reference, and GPT-4 serving as the evaluator. As shown in \Cref{tab:llm_size_coverage}, we found that increasing the model size enhances the coverage scores for both Llama-2 and Vicuna-v1.5-16K. This indicates that \textbf{more parameters improve LLM’s ability to identify diverse information}. \looseness=-1

%% file: tables/llm_performance.tex
\begin{table}[t]
    \small
    \centering
    
    \begin{tabularx}{0.5\textwidth}{lcc}
        \toprule
        
        \textbf{Model}   & \textbf{Faithfulness (\%)} & \textbf{Coverage (\%)}\\
        
        \midrule
        \multicolumn{3}{c}{\textit{Extract then summarize}} \\[2ex]

        \gptfour~ & 95.63& \textbf{36.58}\\
        \vicuna~ & 78.42 & 13.36\\
        \midrule 
        \multicolumn{3}{c}{\textit{Directly summarize}} \\[2ex]
        \gptturbolong~ & \textbf{98.44} & 35.66\\
        \longchat~ & 92.49  & 30.04\\
        
        \bottomrule
    \end{tabularx}

    \vspace{-2mm}
    \caption{Performance of different LLMs on our task. The faithfulness score and coverage score are determined by averaging the binary ratings provided by human evaluators.} %
    
    \label{tab:llm_performance_human}
    \vspace{-5mm}
\end{table}

%% file: tables/position_bias_analysis.tex
\begin{table}[t]
    \small
    \centering
    \begin{adjustbox}{max width=0.48\textwidth}
    {
    \begin{tabular}{lcc|c}
        \toprule
        
        \textbf{Aspect}  & \textbf{First (\%)}   & \textbf{Second (\%)} & \textbf{Consistency (\%)}\\
           
           \midrule
        
        Coverage   & 1.63 &  \textbf{17.55} & 60.10\\
        Faithfulness   & 1.32 & \textbf{13.27} & 61.94\\

        \bottomrule
    \end{tabular}
    }
    \end{adjustbox}
    \vspace{-2mm}
    \caption{Position bias analysis of swapping two summaries produced by two systems. Consistency is calculated as the percentage of cases in which the evaluator (i.e., \gptfour~) provides coherent outcomes upon swapping the order of two summaries. First/Second indicates the percentage of cases in which a judge demonstrates a preference for the first/second summary. Overall, \gptfour~ prefers the summary placed in the second position.} 
    \label{tab:position_bias_analysis}
\end{table}

%% file: tables/verbosity_bias_analysis.tex
\begin{table}[t]
    \small
    \centering
    \begin{adjustbox}{max width=0.48\textwidth}
    {
    \begin{tabular}{lccc}
        \toprule
        
        \textbf{Aspect} & \textbf{Protocol}   & \textbf{Original (\%)}   & \textbf{Extended (\%)} \\
           
           \midrule
        
        \multirow{2}{*}{Faithfulness} & Single  & 41.44 &  20.58\\
        & Pairwise  & \textbf{0.20} &  \textbf{0.00} \\
         
        \midrule
        \multirow{2}{*}{Coverage}  & Single  & 53.46 &  16.33\\ 
        & Pairwise  & \textbf{1.12} &  \textbf{0.82} \\

        \bottomrule
    \end{tabular}
    }
    \end{adjustbox}
    \vspace{-2mm}
    \caption{Verbosity bias analysis using \gptfour~ as the evaluator. Single (i.e., single-answer grading) results in significant verbosity bias as we can see shorter summaries (i.e., Original) are preferable to longer summaries (i.e., Extended). Such bias can be significantly mitigated if pairwise comparison is used instead.} 
    \label{tab:verbosity_bias_analysis}
    \vspace{-5mm}
\end{table}

%% file: tables/protocol_correlation.tex
\begin{table*}[t]
    \small
    \centering
    \begin{adjustbox}{max width=0.98\textwidth}
    {
    \begin{tabular}{lcccccc}
        \toprule
        
        \textbf{Criteria} & \textbf{Reference}  & \textbf{Evaluated Texts} & \textbf{Rating Method}  & \textbf{Evaluator} & \textbf{Rating} & \textbf{Correlation (\%)}\\
           
           \midrule
        
        \multirow{7}{*}{Faithfulness} &  Article &  Summaries &  Pairwise (both ways) &  \gptfour~ &  Win-Tie-Lose &  \textbf{26.68} \\
         &  \CC{50} Article & \CC{50} Summary & \CC{50} Single-answer grading & \CC{50} \gptfour~ & \CC{50} Likert & \CC{50} \underline{21.18}\\
        & Article & Summary & Single-answer grading & \gptfour~ & Binary & 18.54\\
        & Articles & Summary & Single-answer grading & GPT-3.5-Turbo-16K & Likert & -7.44\\
        & Articles & Summary & Single-answer grading & GPT-3.5-Turbo-16K & Binary & -3.70\\
        & Articles & Summary sentence & Single-answer grading & GPT-3.5-Turbo-16K & Likert & 15.58 \\
        & Articles & Summary sentence & Single-answer grading & GPT-3.5-Turbo-16K & Binary & -12.30 \\
        
        \midrule

        \multirow{5}{*}{Coverage} & QA pairs & Summaries & Pairwise (both ways) & \gptfour~ & Win-Tie-Lose & 32.00\\
         & \CC{50} QA pairs & \CC{50} Summary & \CC{50} Single-answer grading & \CC{50} \gptfour~ & \CC{50} Likert & \CC{50} \textbf{36.75}~ \\
        & QA pairs & Summary & Single-answer grading & \gptfour~ & Binary &  22.57 \\
        & QA pair & Summary & Single-answer grading & \gptfour~ & Likert & 29.05\\
         & \CC{50} QA pair & \CC{50} Summary & \CC{50} Single-answer grading & \CC{50} \gptfour~ & \CC{50} Binary & \CC{50} \underline{35.83}\\

        \bottomrule
    \end{tabular}
    }
    \end{adjustbox}
    \vspace{-2mm}
    \caption{Summary-level correlation between different LLM-based evaluation protocols and human judgments computed using Kendall's Tau. The best and second best protocol for each criterion are marked in boldface and underlined, respectively. The recommended evaluation protocols are \hlc{lightblue}{highlighted}.} 
    \label{tab:protocol_correlation}
\end{table*}

%% file: tables/qualitative_examples_qa.tex
\begin{table*}[t]
    \small
    \centering
    \begin{adjustbox}{max width=0.98\textwidth}
    {
    \begin{tabular}{p{0.4\linewidth} p{0.55\linewidth}}
        \toprule
        
        \textbf{Reference} & \textbf{Generated Summary} \\
           
           \midrule
        
        \textbf{Question}: Why do some people welcome the idea of "pumping the brakes" on AI development? \newline
        \textbf{Answers}: 
        \begin{itemize}[leftmargin=*,noitemsep,nolistsep]
            \item[{\color{normalgreen} \cmark~}] bad actors will use the technology improperly
            \item[{\color{normalgreen} \cmark~}] the need for safety limits and regulation of AI tools 
        \end{itemize}& 
        OpenAI CEO Sam Altman believes that artificial intelligence technology, such as the company's AI language model ChatGPT, has the potential to reshape society and improve our lives significantly. However, Altman also expresses real concerns about the dangers and misuse of AI. He worries that AI could be used for large-scale disinformation and offensive cyberattacks. Altman also highlights the importance of involving regulators and society in the development and regulation of AI systems... \\

        \midrule
        
        \textbf{Question}: What are the broader implications of insulin prices on the overall healthcare system? \newline
        \textbf{Answers}: 
        \begin{itemize}[noitemsep,nolistsep,leftmargin=*]
            \item[{\color{darkred} \xmark~}] ..., there is no guarantee that companies will continue to choose people over profits. \looseness=-1
            \item[{\color{darkred} \xmark~}] Legally capping the price of insulin would bring a welcome financial relief to millions of Americans and save lives.
            \item[{\color{darkred} \xmark~}] ... policies to make diabetes treatment more affordable. \looseness=-1
            \item[{\color{darkred} \xmark~}] ... it can hurt patients on fixed incomes and others less able to afford their insulin.
        \end{itemize}& 
        Multiple pharmaceutical companies, including Eli Lilly, Novo Nordisk, and Sanofi, have announced price reductions for their insulin products in response to growing pressure to lower the cost of this life-saving medication. The reductions range from 70\% to 78\% and are aimed at both Medicare patients and individuals with private insurance. The price cuts come as a result of the Inflation Reduction Act, which caps insulin prices for Medicare patients at \$35 per month. However, some critics argue that these reductions only cover a portion of the insulin market and that individuals without insurance or with high-deductible plans may still struggle to afford the medication...\\ %

        \bottomrule
    \end{tabular}
    }
    \end{adjustbox}
    \vspace{-2mm}
    \caption{Two instances in our \datashort~ dataset and corresponding summaries generated by GPT-3.5-Turbo-16K. References and summaries are truncated due to space limits. The references in these two examples contain different types of questions. In the first instance, GPT-3.5-Turbo-16K successfully identifies the answers, demonstrating its proficient comprehension skills. However, in the second instance, the model fails to provide a high-coverage summary. This likely signifies its struggle with complex reasoning tasks that certain types of questions demand.  \looseness=-1} 
    \label{tab:qualitative_examples_qa}
    \vspace{-2mm}
\end{table*}

%% file: tables/size_coverage.tex
\begin{table}[t]
    \small
    \centering
    \begin{adjustbox}{max width=0.48\textwidth}
    {
    \begin{tabular}{lrc}
    
        \toprule
        
        \textbf{Model}   & \textbf{Size} & \textbf{Coverage Score}\\
        
        \midrule

        Llama-2 & 7B & 2.29\\
        Llama-2 & 13B & 2.53\\
        Llama-2 & 70B & 2.81 \\
        \midrule 
        
        Vicuna-v1.5-16K & 7B & 2.00\\
        Vicuna-v1.5-16K & 13B  & 2.02\\
        
        \bottomrule
    \end{tabular}
    }
    \end{adjustbox}
    \vspace{-2mm}
    \caption{Coverage score with regard to LLMs of varying sizes. The coverage scores are computed using the single-answer Likert-scale evaluation protocol with question-and-answer pairs as the reference.} 
    \label{tab:llm_size_coverage}
    \vspace{-5mm}
\end{table}

%% file: contents/02_related_work.tex
\section{Related Work}

\subsection{Multi-document Summarization}
Conventional approaches to multi-document summarization (MDS) can be categorized into three types: extractive \cite{radev-etal-2000-centroid, gillick-favre-2009-scalable, lin-bilmes-2011-class, hong-nenkova-2014-improving, peyrard-eckle-kohler-2016-general, cheng-lapata-2016-neural, narayan-etal-2018-ranking, DBLP:journals/corr/abs-1801-10198}, abstractive \cite{McKeown1995GeneratingSO, radev-mckeown-1998-generating, barzilay-etal-1999-information, DBLP:journals/corr/abs-1804-09010, fabbri-etal-2019-multi}, and multi-sentence compression \cite{ganesan-etal-2010-opinosis, DBLP:conf/ijcai/BanerjeeMS15, chali-etal-2017-towards, nayeem-etal-2018-abstractive}. \looseness=-1

Recently, large language models (LLMs) have demonstrated significant advantages over conventional approaches in generating summaries of high faithfulness and quality. Studies have used LLMs to generate summaries of multiple documents by first extract important sentences from each article and then summarize them \cite{bhaskar-etal-2023-prompted} or iteratively improve summary quality with the guidance of a checklist \cite{DBLP:journals/corr/abs-2305-14647}.

\subsection{MDS Datasets}

In previous studies, several popular MDS datasets have been examined. These datasets include \textsc{DUC} \cite{over2004introduction, dang2005overview} %
and \textsc{TAC} \cite{dang2008overview, owczarzak2011overview}, %
which are smaller in scale with approximately 50 and 100 article clusters, respectively. \textsc{MultiNews} \cite{fabbri-etal-2019-multi} is the first large-scale MDS dataset in the news domain, containing 56K article clusters, with an average of fewer than 3 news articles per cluster. \textsc{auto-hMDS} \cite{zopf-2018-auto} is a multi-lingual MDS dataset focused on the Wikipedia domain, comprising 7.3K article clusters. \textsc{WCEP} \cite{gholipour-ghalandari-etal-2020-large} is another Wikipedia domain dataset, where each cluster may contain up to 100 articles. \textsc{Multi-XScience} \cite{lu-etal-2020-multi-xscience} and \textsc{MS\textasciicircum2} \cite{deyoung-etal-2021-ms} are two scientific domain MDS datasets. The above MDS datasets task models with summarizing consensus information, our work differentiates itself by focusing on summarizing diverse information across the articles.

%% file: contents/06_conclusion.tex
\section{Conclusion}

We introduce a novel task of Multi-document Diverse Summarization that focuses on effectively summarizing diverse information from multiple news articles discussing the same news story. To facilitate this task, we construct a dataset, \datashort~, using our proposed QA-based pipeline. Through meticulous human evaluation, we have demonstrated that although LLMs exhibit a high level of faithfulness in tackling our task, achieving a high coverage rate remains particularly challenging, even with the most advanced LLM like GPT-4. This underscores both the challenges and opportunities of \taskshort~. 

Furthermore, we have conducted an extensive analysis of bias and its correlation with human assessments across a range of evaluation protocols. Leveraging the insights obtained from these experiments, we propose a set of recommendations that outline the most effective protocols for evaluating model performance within our task domain. Our paper also delves into a comprehensive study that investigates LLMs' tendency to summarize various types of information. The outcomes of these analyses offer valuable insights into the behaviors exhibited by different LLMs when they engage with the challenge of summarizing diverse information. By presenting these resources and research findings, we hope to inspire and motivate future endeavors in the realm of comprehending and summarizing the intricate nuances present in diverse news articles concerning the same news event.

%% file: contents/07_ethics.tex
\section{Ethical Considerations}
In \Cref{sec:data} and \Cref{subsec:rq1}, we engaged annotators for data annotation and human evaluation. We prioritized fair compensation for our participants, with details provided in \Cref{apx:rq2_discussion}. To foster an ethical working environment, we allowed participants to set their own pace, facilitated open communication for any concerns, and provided the option to withdraw from the project at any time without repercussions. Additionally, we took measures to ensure the anonymity of the data annotations by avoiding the inclusion of any personally identifiable information. \looseness=-1

%% file: contents/08_limitation.tex
\section{Limitation}

This study contributes significantly to the field of multi-document summarization by providing a larger and more comprehensive dataset than those available in previous research. However, there are several limitations that must be acknowledged.

Firstly, despite our best efforts to curate a large enough dataset, it still represents a relatively small fraction of the vast array of news content available online. This limitation is intrinsic to the task at hand, given the financial implications of human annotation and the complexity of multi-document summarization necessitates that annotators thoroughly read and understand multiple articles, which exponentially increases the time and cost associated with the annotation process compared to single-document summarization.

Moreover, while we carried out thorough LLM-based evaluations, we did not investigate the exact influence of different prompts on the LLM's performance. Even though we have tried our best to manually optimize the prompts, the lack of analysis on prompt sensitivity could lead to slightly different outcomes.

Furthermore, as our dataset encompasses online news articles, the study may not adequately capture the complexity of summarizing documents from diverse domains. News articles often follow a particular structure, which might not be prevalent in other kinds of multi-document contexts, such as academic papers or legal documents. Consequently, the generalizability of our findings and the utility of the dataset beyond the news domain demands further analysis.

%% file: contents/appendix.tex
\clearpage
\appendix

\section{Are our findings in \Cref{subsec:req2} still reproducible after a GPT-4 update every two months? } 
\label{apx:rq2_discussion}
While it's a valid concern that the evolution of GPT models could impact the reproducibility of our findings, it's important to note that the principles highlighted in this research are not necessarily tied to the specific version of the GPT model itself, but rather how these language models work conceptually. The potential biases and evaluation techniques of GPT-4 we discuss can likely be applied or adapted to newer versions as well. 

Naturally, with the release of an updated model, a new set of tests would be ideal to validate whether these findings hold. But this is true of any research in changing and evolving fields and does not detract from the value of our current findings. If anything, our research forms a foundation to more effectively assess future iterations of the GPT models in terms of evaluating coverage and faithfulness.
\section{Human Annotation}
In this section, we illustrate the details of our human annotation process.
\begin{figure*}[b]
    \centering
    \includegraphics[width=0.95\linewidth]{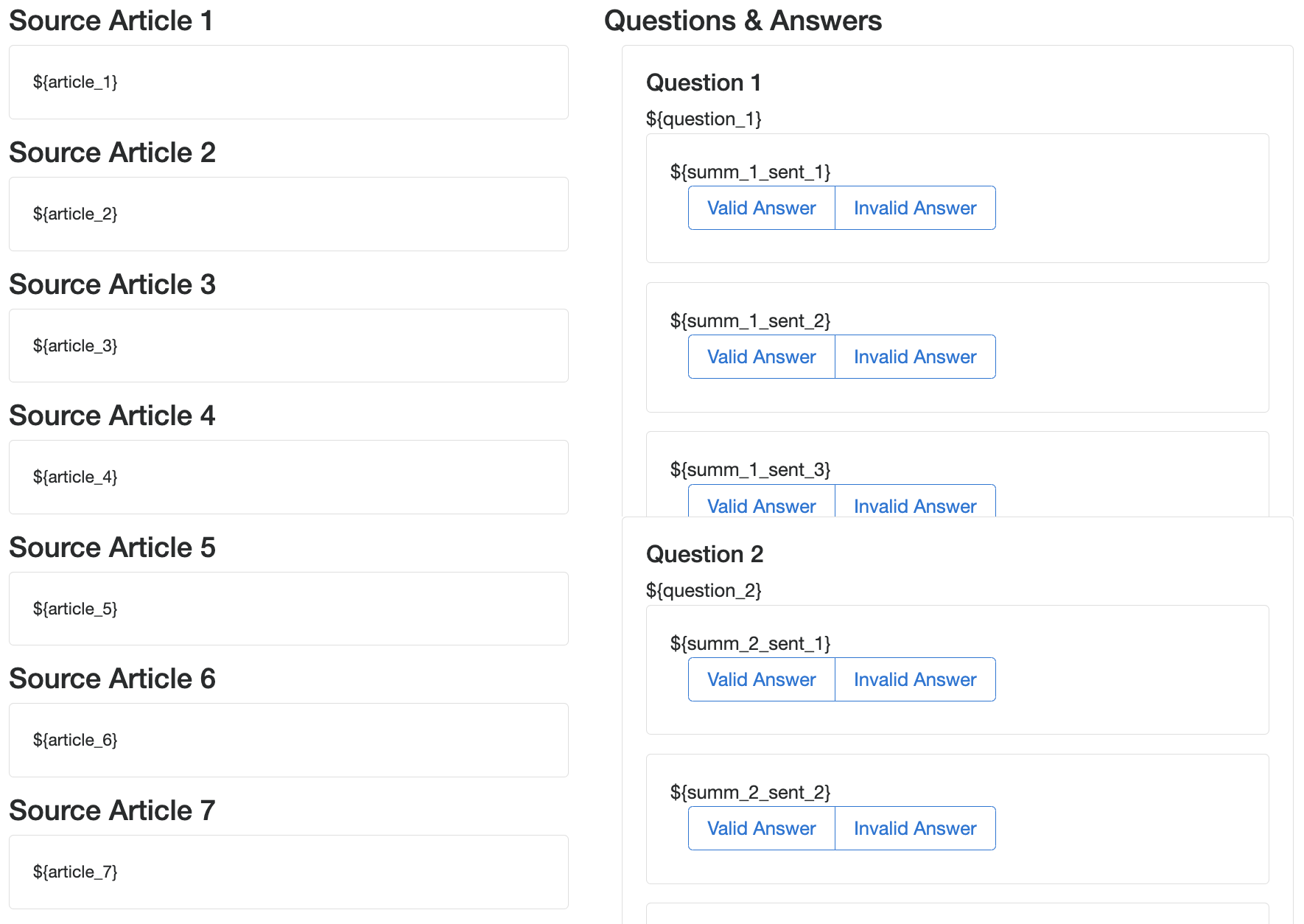}
    \vspace{-2mm}
    \caption{Annotation interface for filtering invalid QA pairs.} 
    \label{fig:interface_filter_qa}
    \vspace{-5mm}
\end{figure*}

\subsection{Worker Qualification}
We established specific preliminary criteria for the recruitment of MTurk workers who possess strong performance histories. These criteria include having a HIT approval rate of 99\% or higher, having approved a minimum of 10,000 HITs, and being located within the United Kingdom, Canada, and the United States. 

Furthermore, apart from these preliminary criteria, eligible workers are required to pass three rounds of qualification tests centered around the faithfulness evaluation task, which is illustrated in \Cref{tab:llm_performance_human}. To streamline the qualification process, the authors manually annotate 3 HITs. Each HIT comprises ten news articles and four summaries generated by four different models. During each qualification round, annotators are presented with one of these annotated samples. Workers whose annotations do not exhibit a sufficiently high level of agreement with our annotations are excluded from the selection process.

Ultimately, 16 annotators who successfully passed all three rounds of qualification tests were selected. All the human evaluations and annotations are conducted by these 16 annotators. Additionally, every HIT has been meticulously designed to ensure that annotators can achieve an equivalent hourly pay rate of \$20 provided they work continuously.

\subsection{Annotating QAs}
\label{apx:human_annotation_qas}
When annotating QA pairs, annotators are presented with the post-processed results detailed in \Cref{subsec:automatic_data_collection}. Below, we show the guidelines and the annotation interface presented to the annotators...

\paragraph{Guideline}
In this task, you will evaluate the validity of several answers with regard to the corresponding questions.
To correctly solve this task, follow these steps:

\begin{itemize}[noitemsep,nolistsep]
    \item Carefully read the questions, answers, and the source articles.
    \item For each answer, check it against the question and the list of source articles.
    \item An answer is \textbf{Valid} if and only if (1) it \textbf{addresses the question, AND} (2) \textbf{at least one article contains such information} (It does not have to be word by word. It is sufficient that the information presented in the answer can be found in at least one article).

\end{itemize}

\textbf{Warning}: Annotations will be checked for quality against control labels, \textbf{low-quality work will be rejected}.

\textbf{Valid answer}: The validity depends on if the information in the answer is mentioned/supported by any source articles, not if the exact words are stated in the source articles. A valid answer should also provide a response that addresses the question it is paired with.
Answer not addressing the question or suggesting no information should be marked as \textbf{Invalid Answer}. Examples of \textbf{Invalid Answer} are shown below:
\begin{itemize}[noitemsep,nolistsep]
    \item Question: What are the foreign impact of ...? Answer: The domestic influence of ...
    \item The article does not provide a clear answer to ...    
    \item ... is not discussed in the article.
    \item As a language model, I cannot ...
\end{itemize}

\begin{figure*}[b]
    \centering
    \includegraphics[width=0.95\linewidth]{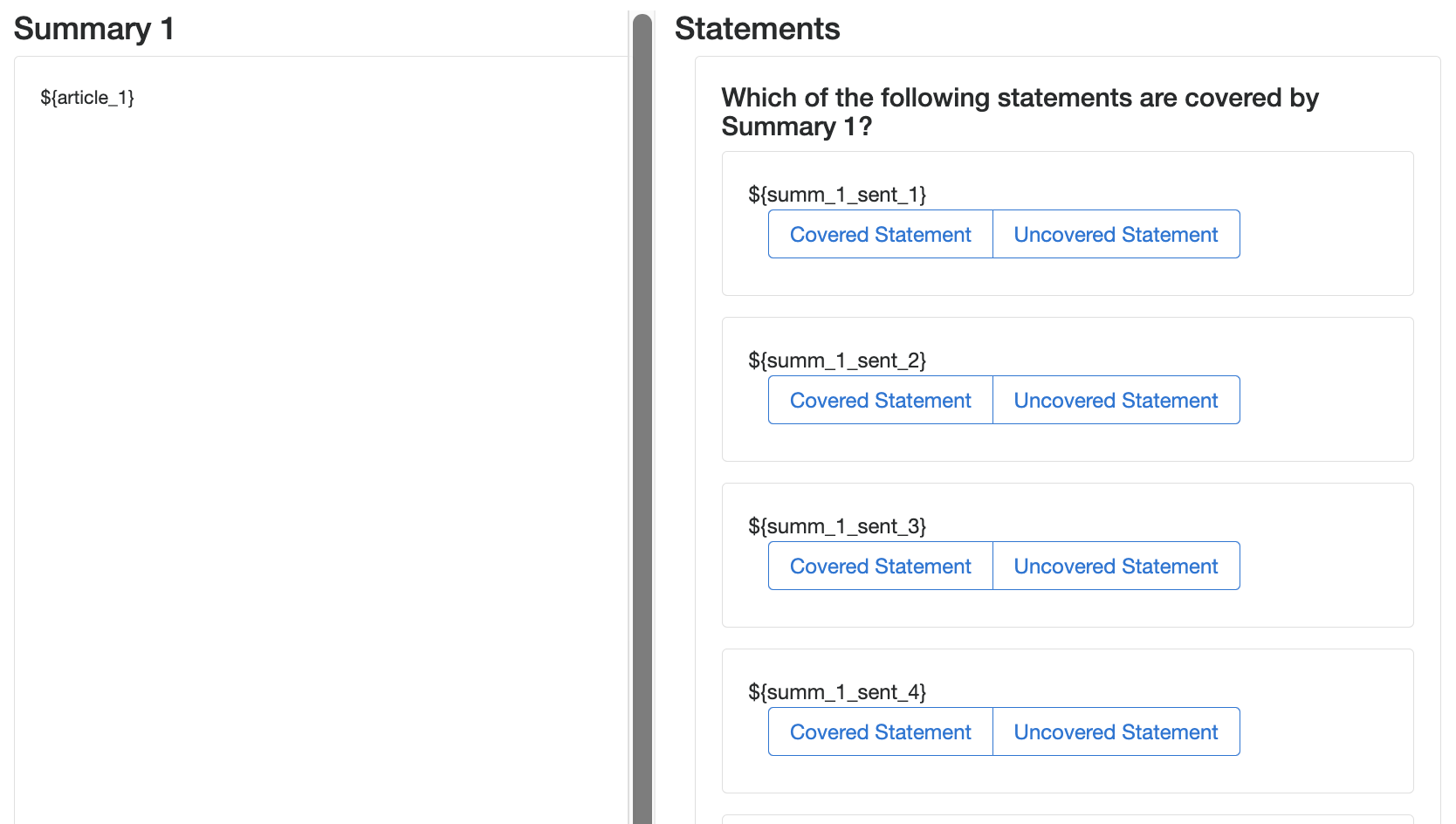}
    \vspace{-2mm}
    \caption{Interface for coverage evaluation.} 
    \label{fig:interface_coverage_eval}
    \vspace{-5mm}
\end{figure*}

\paragraph{Interface}
The annotation interface for filtering invalid QA pairs is presented in \Cref{fig:interface_filter_qa}.

\subsection{Coverage Evaluation}
\label{apx:human_eval}

\paragraph{Guideline}
In this task, you will evaluate the \textit{coverage} of several statements with regard to the corresponding summaries. The statements are derived from news articles.
To correctly solve this task, follow these steps:

\begin{itemize}
\item Carefully read the statements and the summaries.
\item  For each statement, check it against the corresponding summary.
\item  A statement is \textbf{Covered} if and only if it is mentioned or supported by the corresponding summary. (\textbf{It does not have to be word by word}. It is sufficient that the information presented in the statement can be found in the corresponding summary).

\end{itemize}
\begin{figure*}[b]
    \centering
    \includegraphics[width=0.95\linewidth]{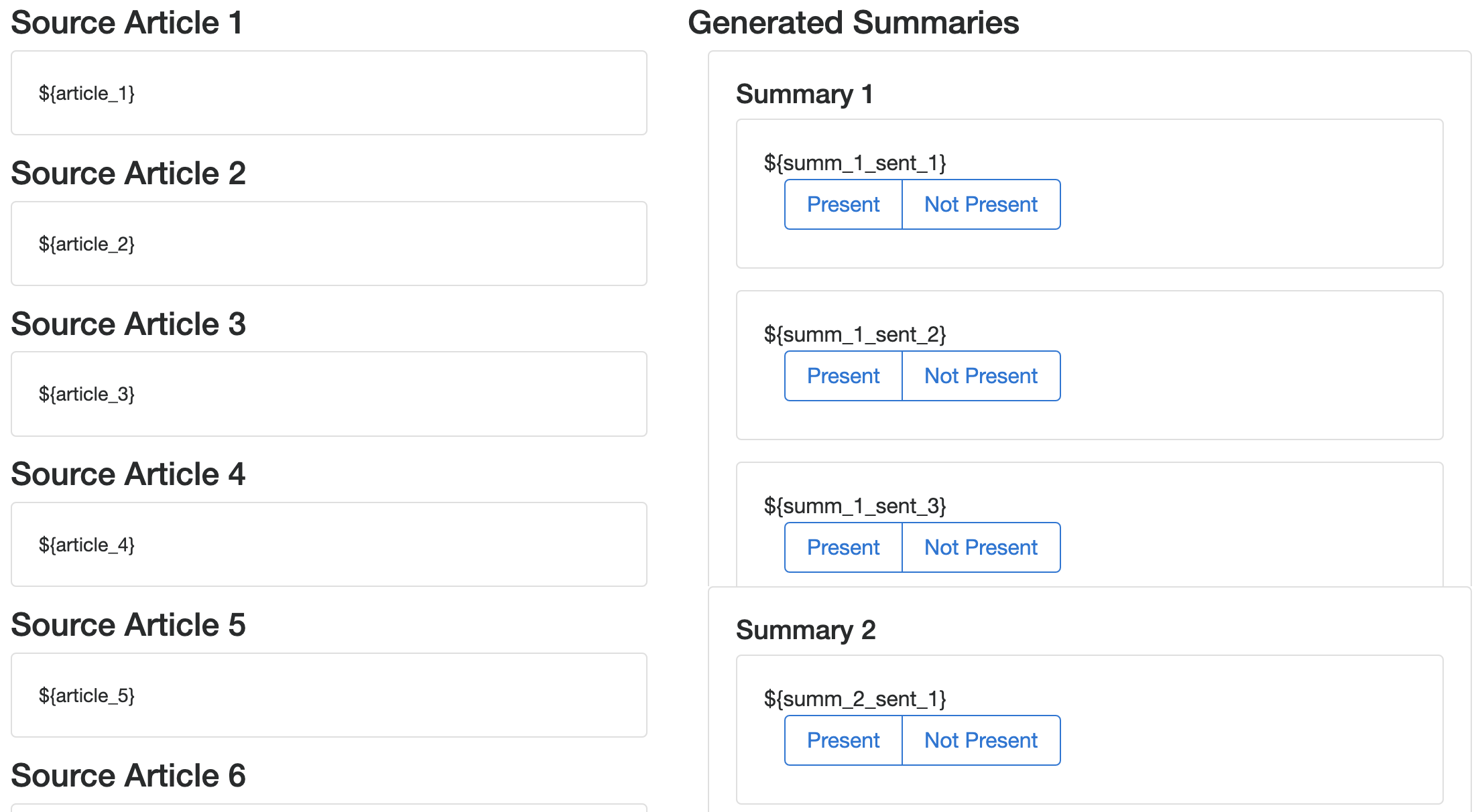}
    \vspace{-2mm}
    \caption{Interface for faithfulness evaluation.} 
    \label{fig:interface_faithfulness_eval}
    \vspace{-5mm}
\end{figure*}
\textbf{Warning}: Annotations will be checked for quality against control labels, \textbf{low-quality work will be rejected}.

\textbf{Covered Statement}: The coverage depends on if the information in the statement is mentioned/supported by the corresponding summary, not if the exact words are stated in the corresponding summary.
Some summaries may contain article number. Please ignore the article number and focus on whether the information in the statement is mentioned/supported by the corresponding summary.

\paragraph{Evaluation Interface}
The interface for coverage evaluation is shown in \Cref{fig:interface_coverage_eval}.

\subsection{Faithfulness Evaluation}

\paragraph{Guidelines}

In this task, you will evaluate the \textit{faithfulness} between each sentence of automatically generated summaries and a list of source articles used to generate the summaries.
To correctly solve this task, follow these steps:

\begin{itemize}
    \item Carefully read the generated summaries and the source articles.
    \item For each sentence, compare it against the list of source articles and decide if any of the source articles support this \textbf{sentence}.
    \item If there is at least one article that supports this sentence, rate the sentence as \textbf{Present}. Otherwise, select \textbf{Not Present}.    
\end{itemize}

\textbf{Warning}: Annotations will be checked for quality against control labels, \textbf{low-quality work will be rejected}.

\textbf{Faithfulness}: The rating depends on if the information in the generated sentence is mentioned/supported by any source articles, not if the exact words are stated in the source articles
Nonsense sentences should always be considered unfaithful, and you should select Not Present. Examples of these are shown below:

\begin{itemize}[noitemsep,nolistsep]
    \item As a language model, I cannot ...
    \item I am ready to summarize...
    \item Please provide the next set of news sentences...
    \item Sentence 1: 1: \textbackslash n* \ n* 1: 1: 1: 1: 1:    
\end{itemize}

\paragraph{Interface}
We display the interface for faithfulness evaluation in \Cref{fig:interface_faithfulness_eval}.

\subsection{Inter-annotator Agreement}
We compute the quality of our annotations and evaluations using Krippendorff's Alpha \cite{krippendorff1970estimating}. For faithfulness and coverage evaluations, the inter-annotator agreement is 0.61 and 0.60, respectively. For reference annotations, the inter-annotator agreement is 0.69. These numbers represent a moderate to high agreement.

\section{LLM Prompts}
In this section, we display all the prompts used in our experiments. Texts marked in boldface indicate placeholders.

\subsection{LLM Prompts for Reference Annotation}
\label{apx:llm_prompts_data}
\begin{figure*}[bt]
    \centering
    \includegraphics[width=0.95\linewidth]{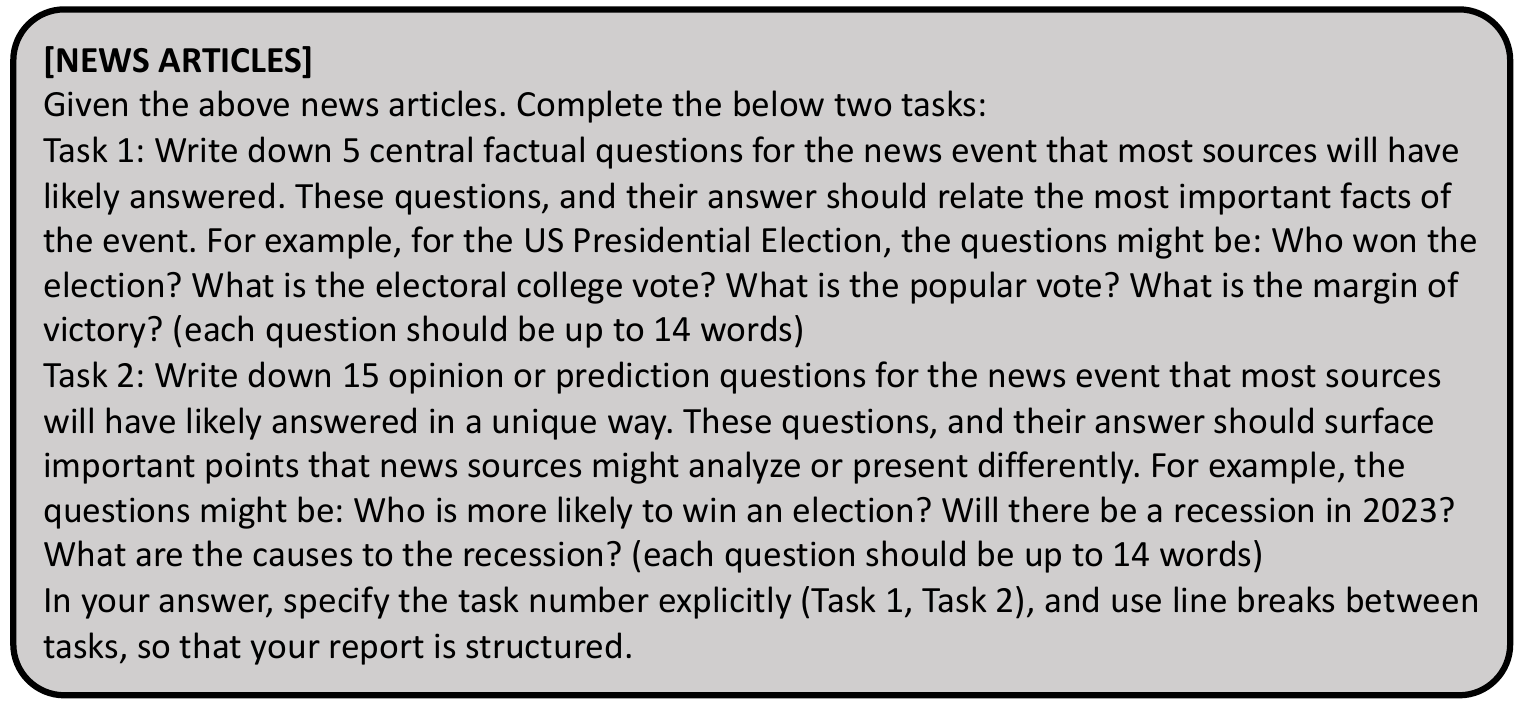}
    \vspace{-2mm}
    \caption{The prompt for question generation.} 
    \label{fig:qg_prompt}
    \vspace{-5mm}
\end{figure*}

\begin{figure*}[bt]
    \centering
    \includegraphics[width=0.95\linewidth]{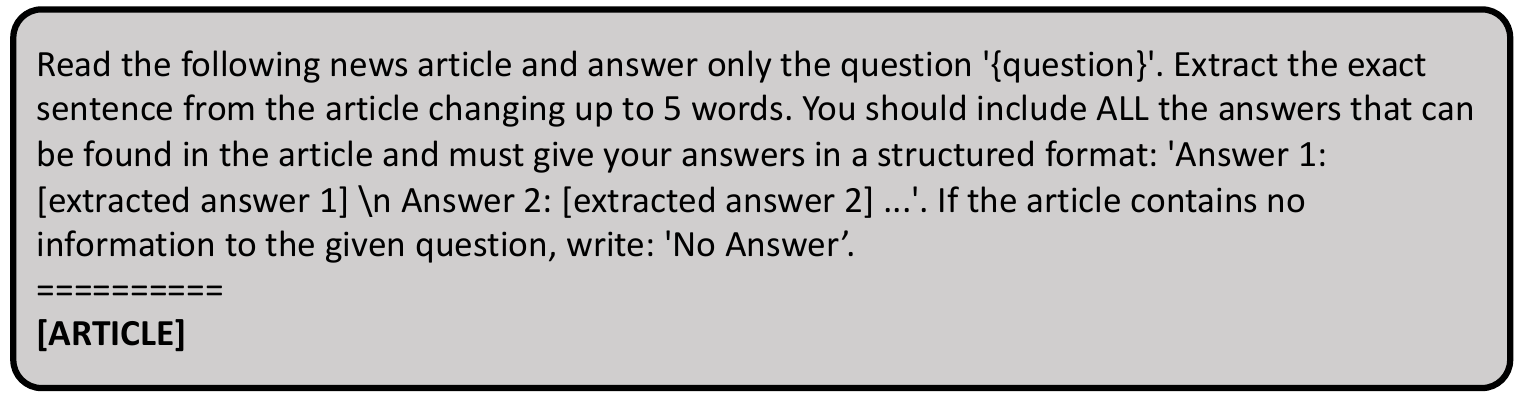}
    \vspace{-2mm}
    \caption{The prompt for question answering.} 
    \label{fig:qa_prompt}
    \vspace{-5mm}
\end{figure*}

\begin{figure*}[bt]
    \centering
    \includegraphics[width=0.95\linewidth]{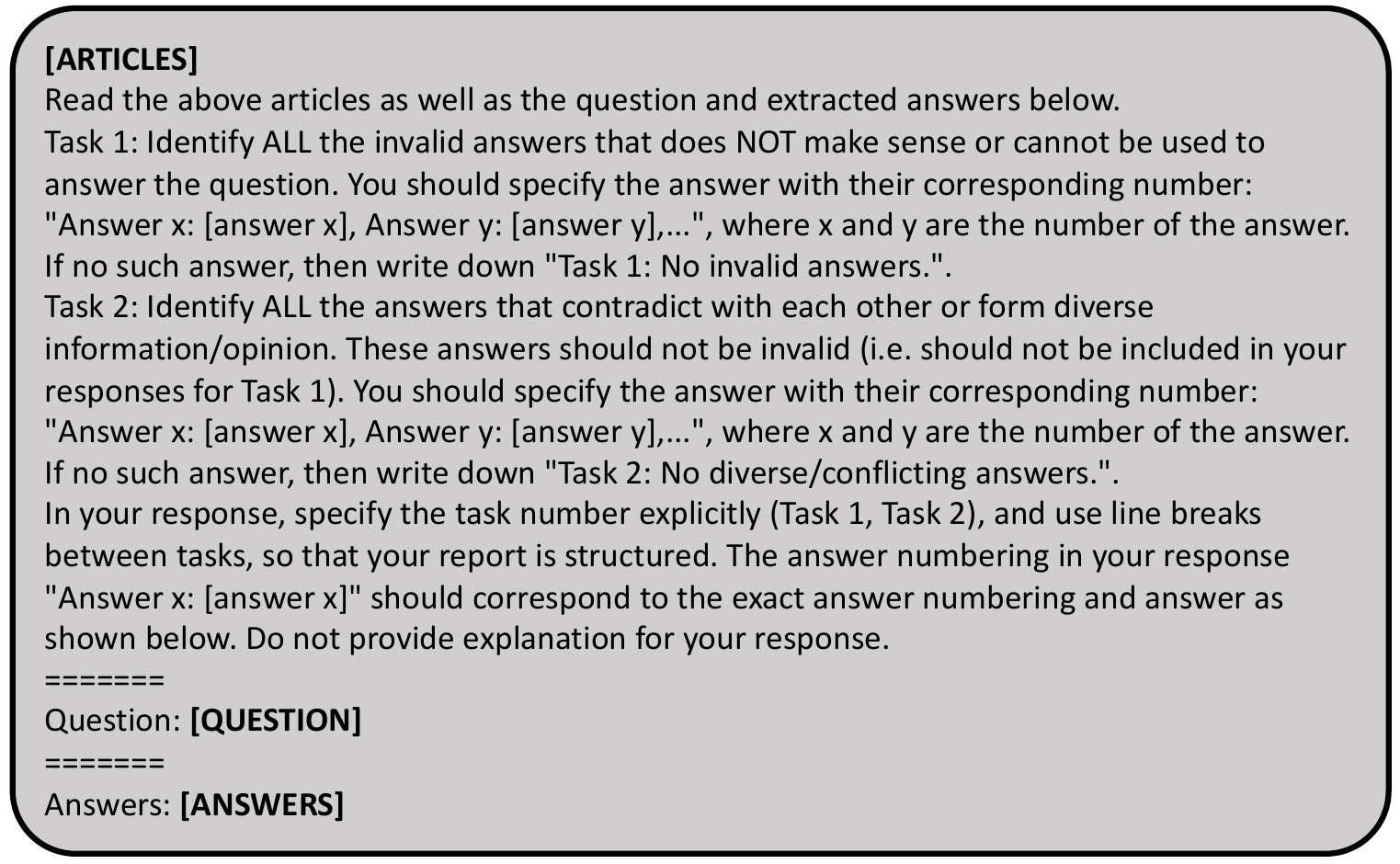}
    \vspace{-2mm}
    \caption{The prompt for post-processing.} 
    \label{fig:postprocessing_prompt}
    \vspace{-5mm}
\end{figure*}

Data collection pipeline consists of three components that are based on prompting ChatGPT: question generation, question answering, and post-processing. The prompt to each component is displayed in \Cref{fig:qg_prompt}, \Cref{fig:qa_prompt}, and \Cref{fig:postprocessing_prompt}, respectively. 

\subsection{LLM Prompts for Summarization}
\label{apx:llm_prompts_summ}
\begin{figure*}[bt]
    \centering
    \includegraphics[width=0.95\linewidth]{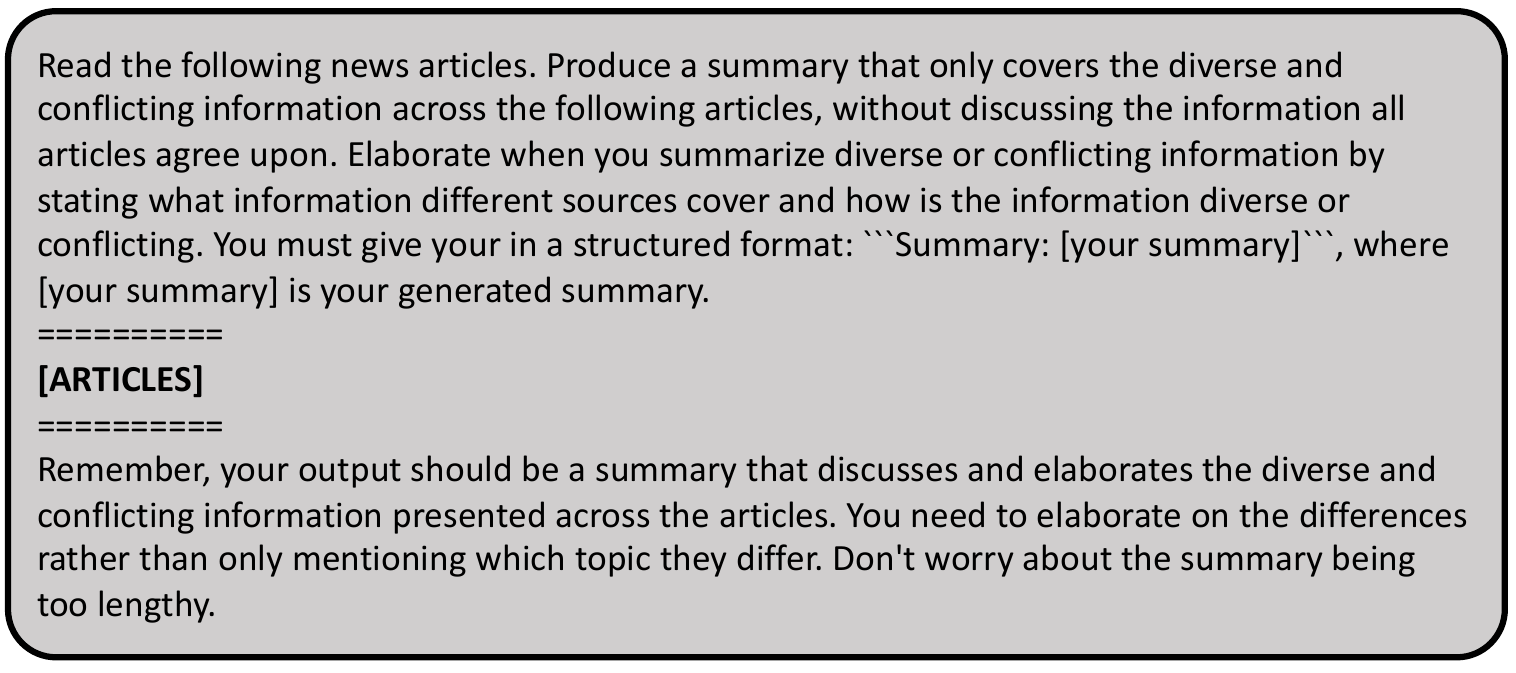}
    \vspace{-2mm}
    \caption{The prompt to long-context LLMs for direct summarization from all input articles.} 
    \label{fig:direct_summ_prompt}
    \vspace{-5mm}
\end{figure*}

\begin{figure*}[bt]
    \centering
    \includegraphics[width=0.95\linewidth]{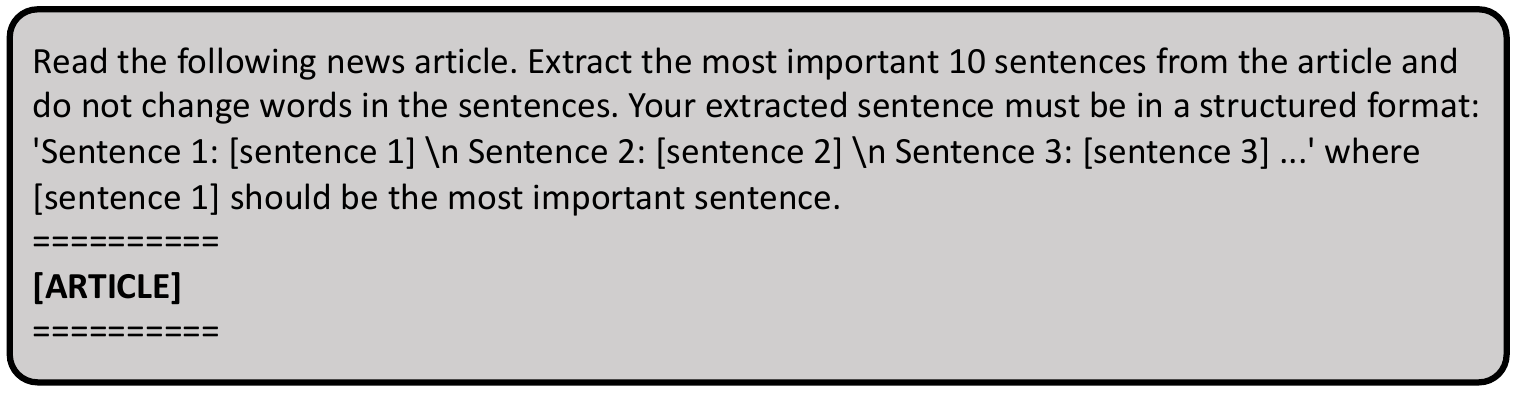}
    \vspace{-2mm}
    \caption{The prompt to standard LLMs for extracting important sentences from a given article.} 
    \label{fig:extract_summ_prompt_1}
    \vspace{-5mm}
\end{figure*}

\begin{figure*}[bt]
    \centering
    \includegraphics[width=0.95\linewidth]{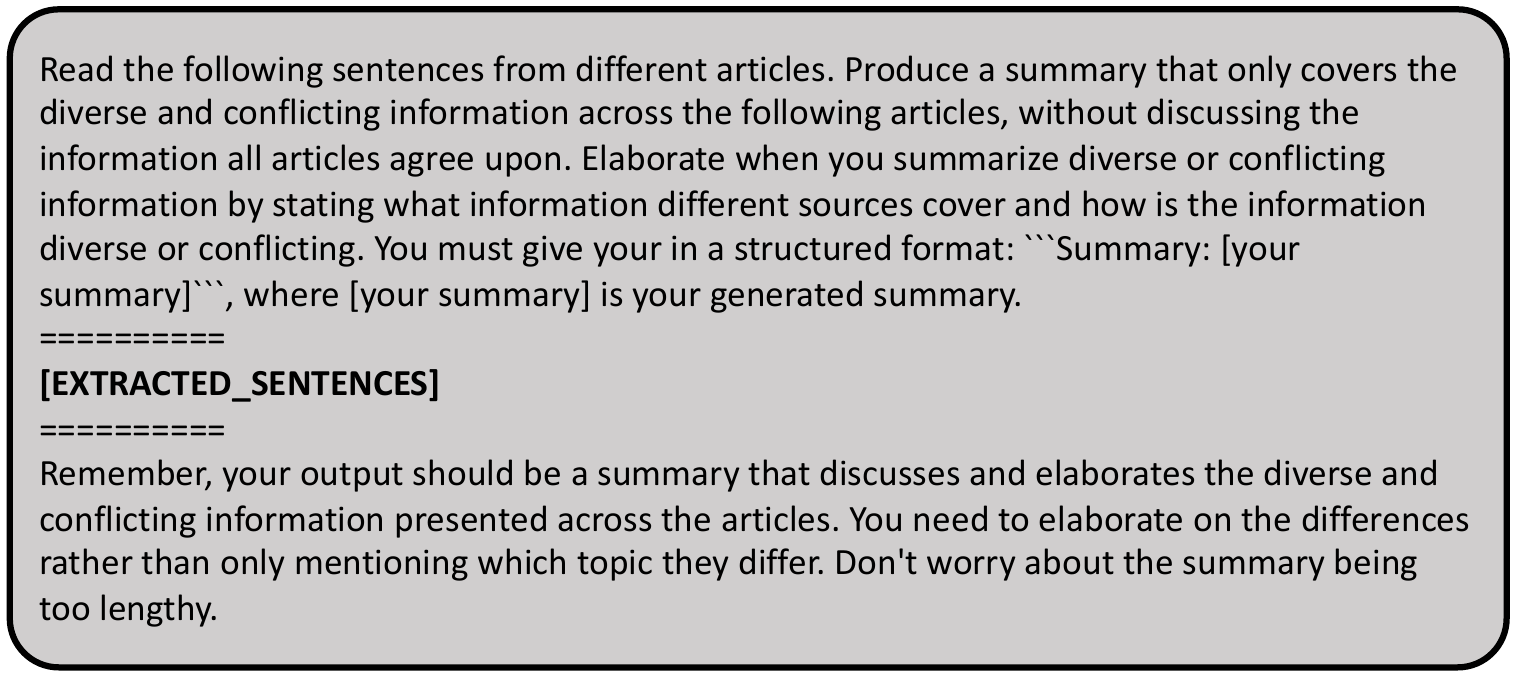}
    \vspace{-2mm}
    \caption{The prompt to standard LLMs for summarizing the extracted sentences.} 
    \label{fig:extract_summ_prompt_2}
    \vspace{-5mm}
\end{figure*}

We use different prompts for long-context and standard LLMs since the latter does not have long enough contexts to process all the input articles. The prompt template for long-context LLMs is displayed in \Cref{fig:direct_summ_prompt}, while the two prompt templates for standard LLMs are shown in \Cref{fig:extract_summ_prompt_1} and \Cref{fig:extract_summ_prompt_2}.

\begin{figure}[bt]
    \centering
    \includegraphics[width=0.95\linewidth]{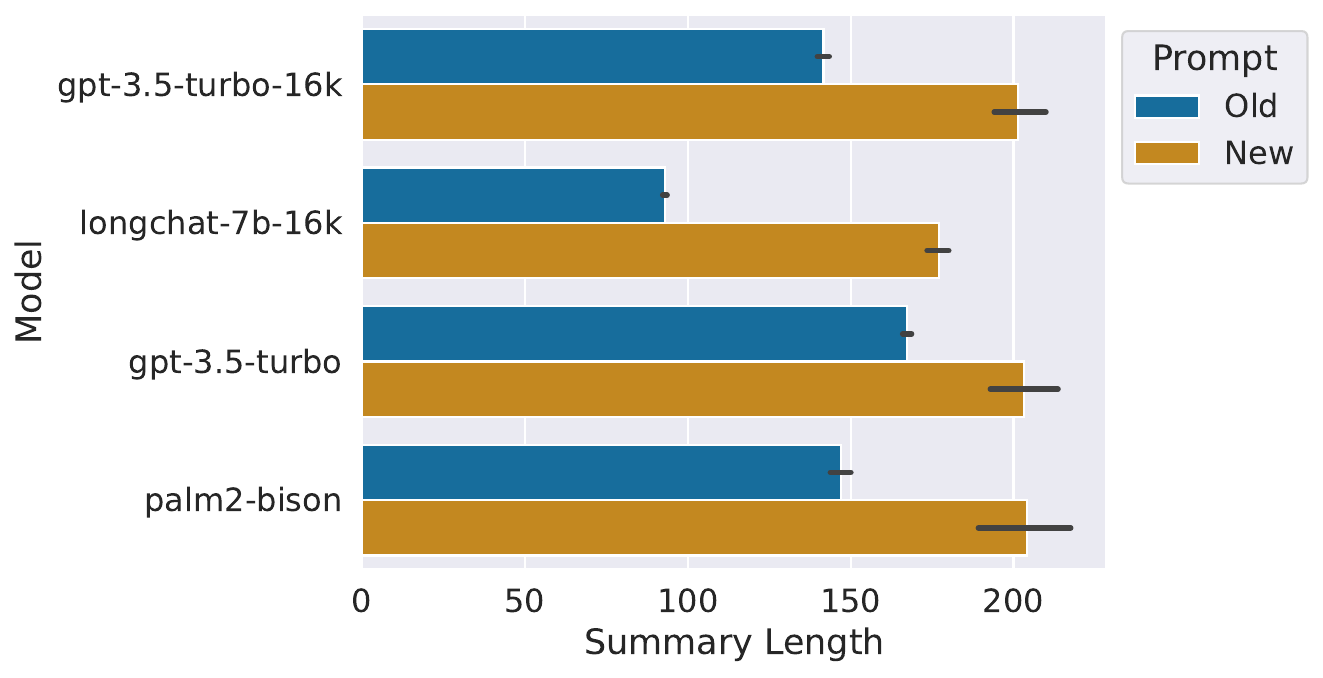}
    \vspace{-2mm}
    \caption{Lengths of summaries (token counts) produced by different models and different prompts. \textbf{New} indicates the final prompt we used, while \textbf{Old} denotes the initial prompt we tried.} 
    \label{fig:summary_length_comparison}
    \vspace{-5mm}
\end{figure}

\begin{figure*}[bt]
    \centering
    \includegraphics[width=0.95\linewidth]{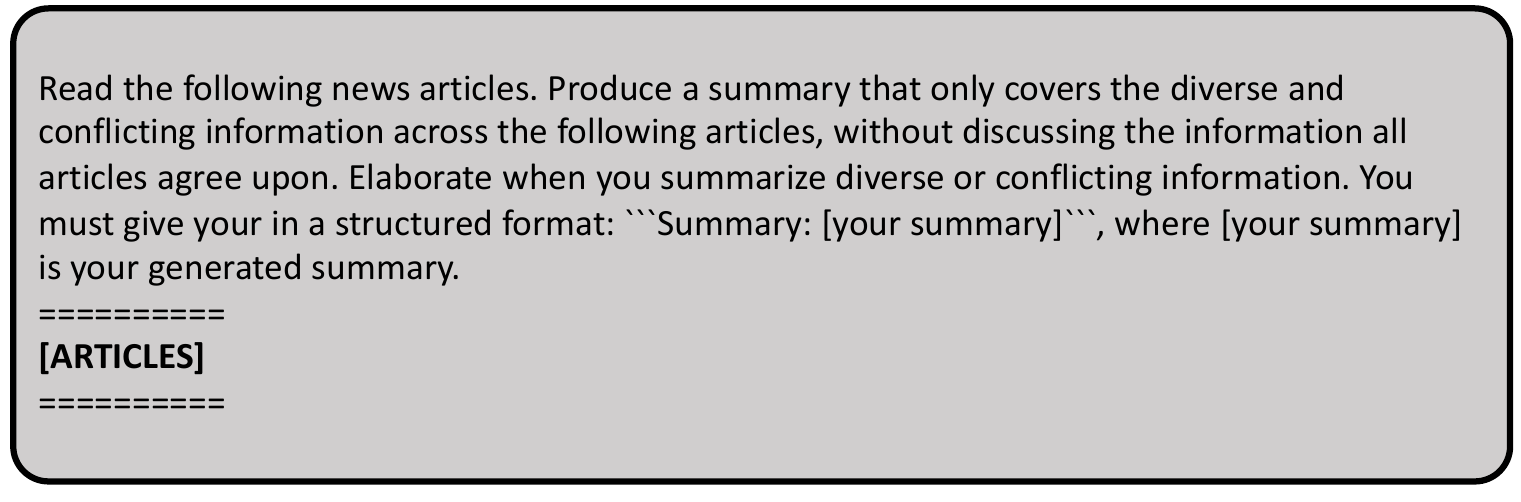}
    \vspace{-2mm}
    \caption{The prompt to standard LLMs for summarizing the extracted sentences.} 
    \label{fig:direct_summ_prompt_initial}
    \vspace{-5mm}
\end{figure*}

Note that the prompts displayed in the above-mentioned figures have undergone meticulous prompt engineering. We found that these prompts in general produce summaries with a higher coverage. In particular, we found that adding ``Don't worry about the summary being too lengthy.'' in the prompt to \gptfour~ is the key to generating more comprehensive summaries. As a comparison, we show our initial prompt to long-context LLMs in \Cref{fig:direct_summ_prompt_initial}, which is much shorter than the prompt in \Cref{fig:direct_summ_prompt}. We use summary length to approximate coverage. As shown in \Cref{fig:summary_length_comparison}, the final prompt we used can significantly increase the length of the generated summaries.

\subsection{LLM Prompts for Evaluation}
\label{apx:llm_prompts_eval}
In this section, we display the prompts to \gptfour~ used in our evaluation.

\begin{figure*}[bt]
    \centering
    \includegraphics[width=0.95\linewidth]{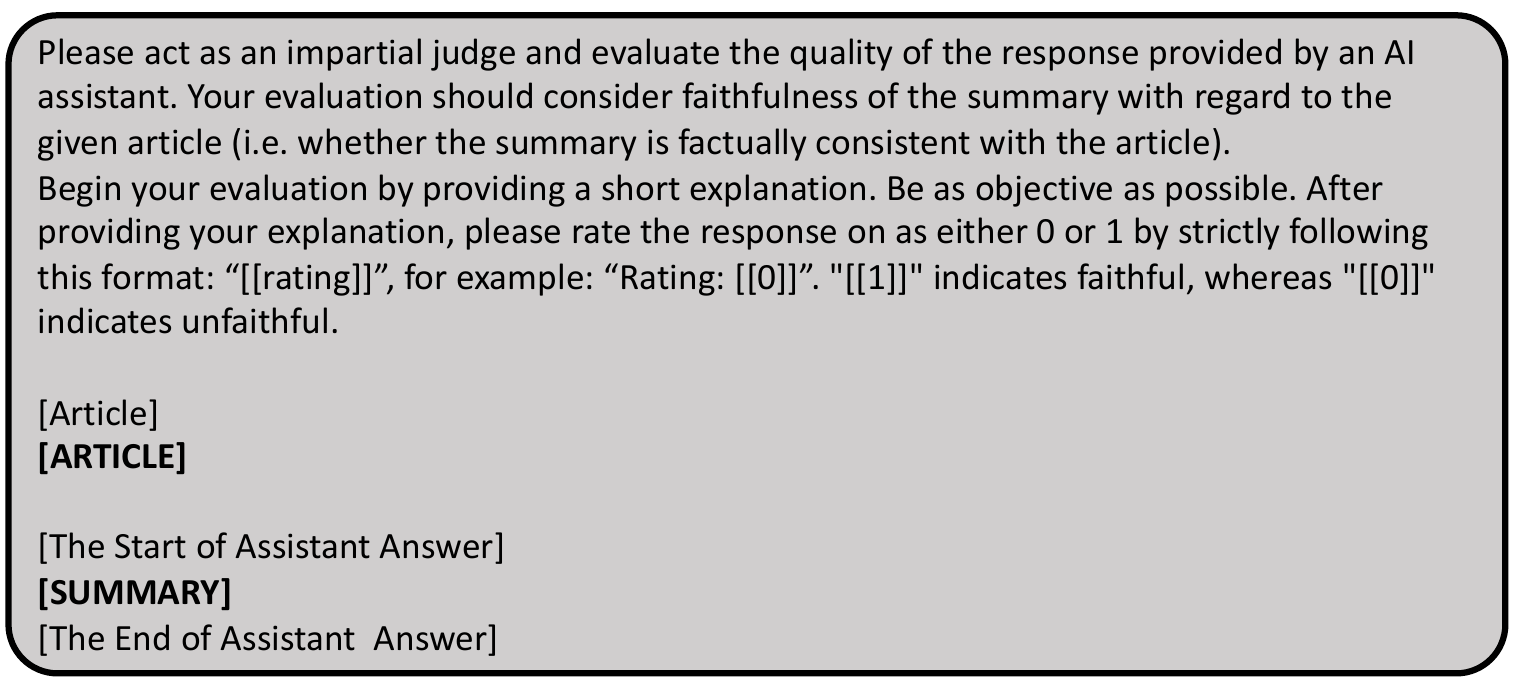}
    \vspace{-2mm}
    \caption{The prompt to \gptfour~ for the binary single-answer grading faithfulness evaluation protocol.} 
    \label{fig:faithfulness_binary}
    \vspace{-5mm}
\end{figure*}

\begin{figure*}[bt]
    \centering
    \includegraphics[width=0.95\linewidth]{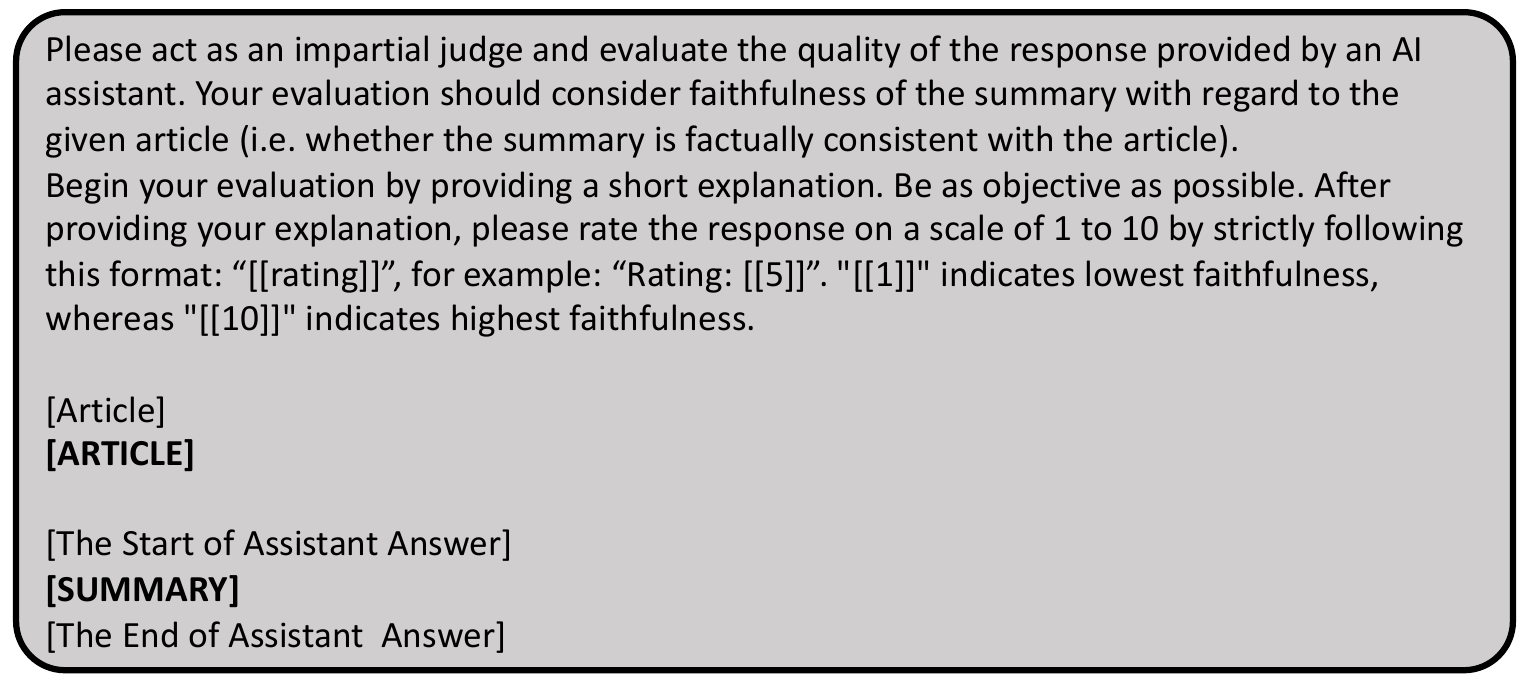}
    \vspace{-2mm}
    \caption{The prompt to \gptfour~ for the Likert-scale single-answer grading faithfulness evaluation protocol.} 
    \label{fig:faithfulness_likert}
    \vspace{-5mm}
\end{figure*}

\begin{figure*}[bt]
    \centering
    \includegraphics[width=0.95\linewidth]{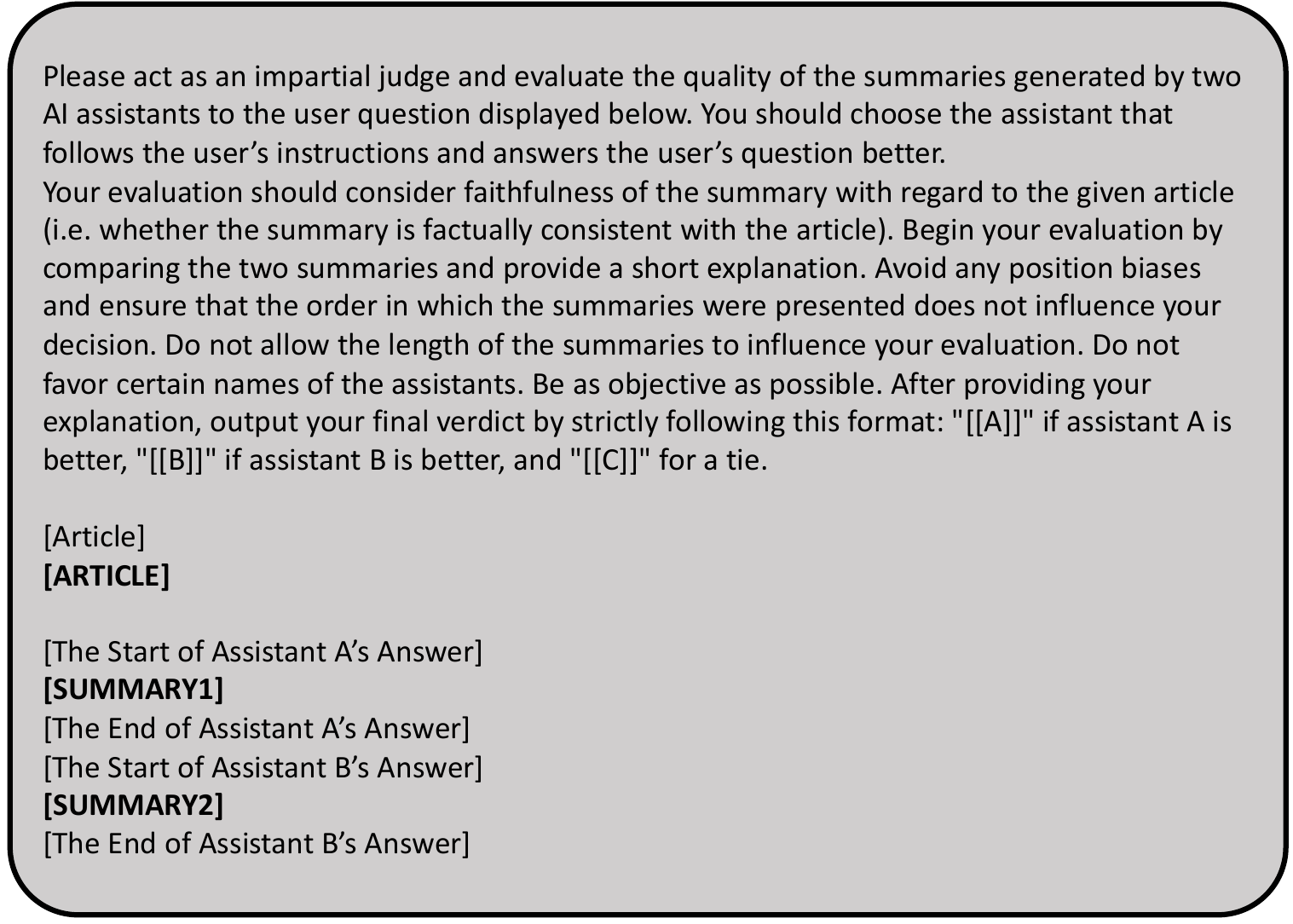}
    \vspace{-2mm}
    \caption{The prompt to \gptfour~ for the pairwise comparison faithfulness evaluation protocol.} 
    \label{fig:faithfulness_pair}
    \vspace{-5mm}
\end{figure*}

\begin{figure*}[bt]
    \centering
    \includegraphics[width=0.95\linewidth]{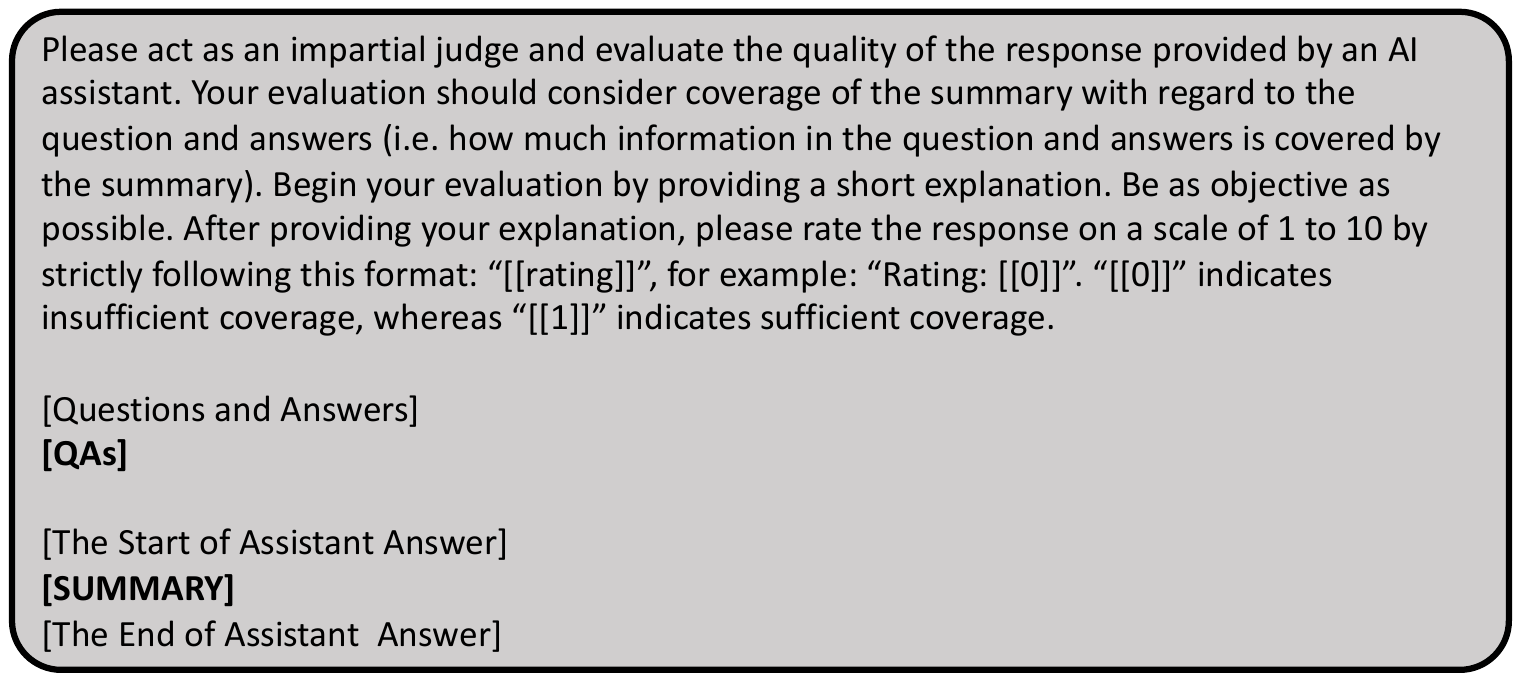}
    \vspace{-2mm}
    \caption{The prompt to \gptfour~ for the binary single-answer grading coverage evaluation protocol.} 
    \label{fig:coverage_binary}
    \vspace{-5mm}
\end{figure*}

\begin{figure*}[bt]
    \centering
    \includegraphics[width=0.95\linewidth]{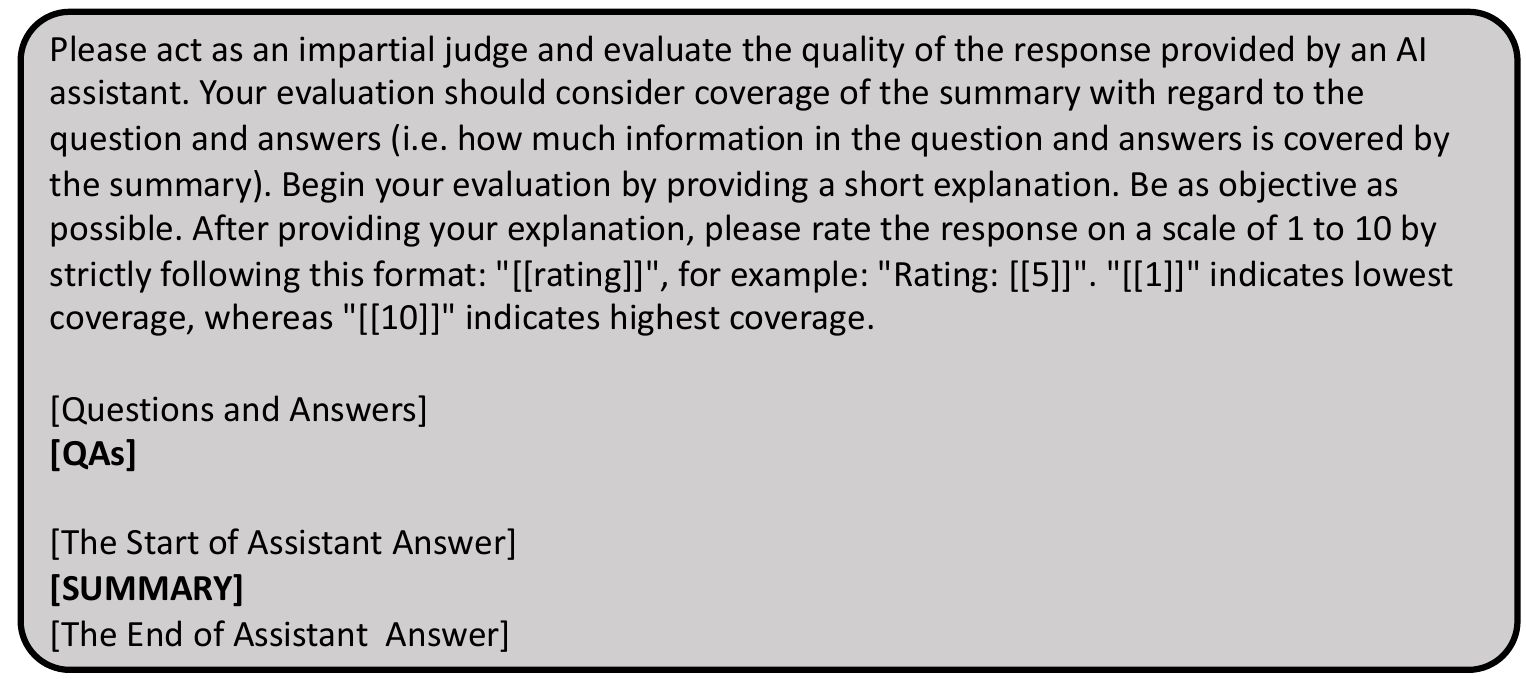}
    \vspace{-2mm}
    \caption{The prompt to \gptfour~ for the Likert-scale single-answer grading coverage evaluation protocol.} 
    \label{fig:coverage_likert}
    \vspace{-5mm}
\end{figure*}

\begin{figure*}[bt]
    \centering
    \includegraphics[width=0.95\linewidth]{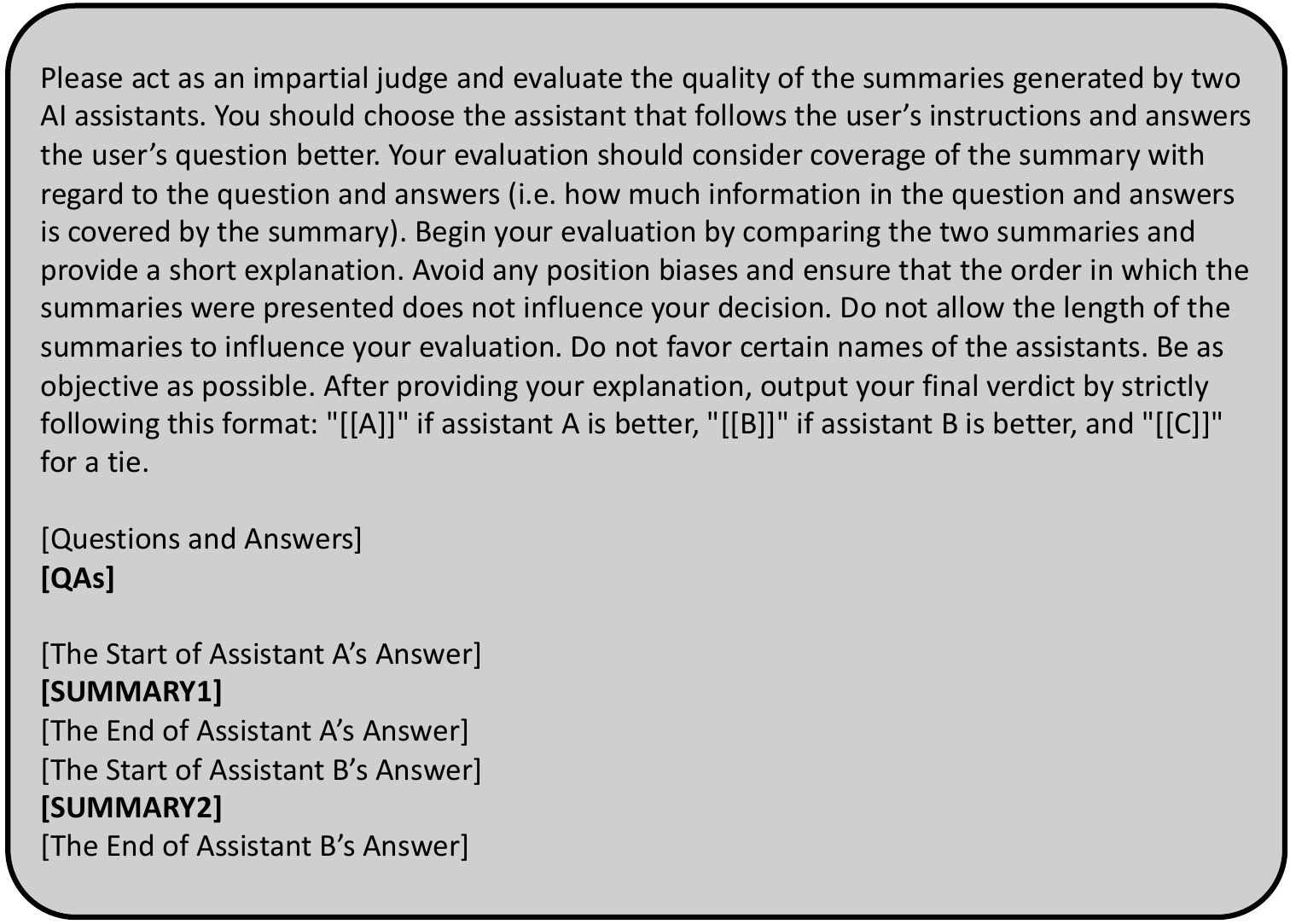}
    \vspace{-2mm}
    \caption{The prompt to \gptfour~ for the pairwise comparison coverage evaluation protocol.} 
    \label{fig:coverage_pair}
    \vspace{-5mm}
\end{figure*}

\section{LLM Bias Analysis}
\label{apx:llm_bias_analysis}

In this section, we present the details of the bias analysis we conducted in \Cref{subsec:req2}.

\subsection{Position Bias}
\begin{figure*}[t]
    \centering
    \begin{subfigure}{0.48\textwidth}

    \includegraphics[width=.95\linewidth]{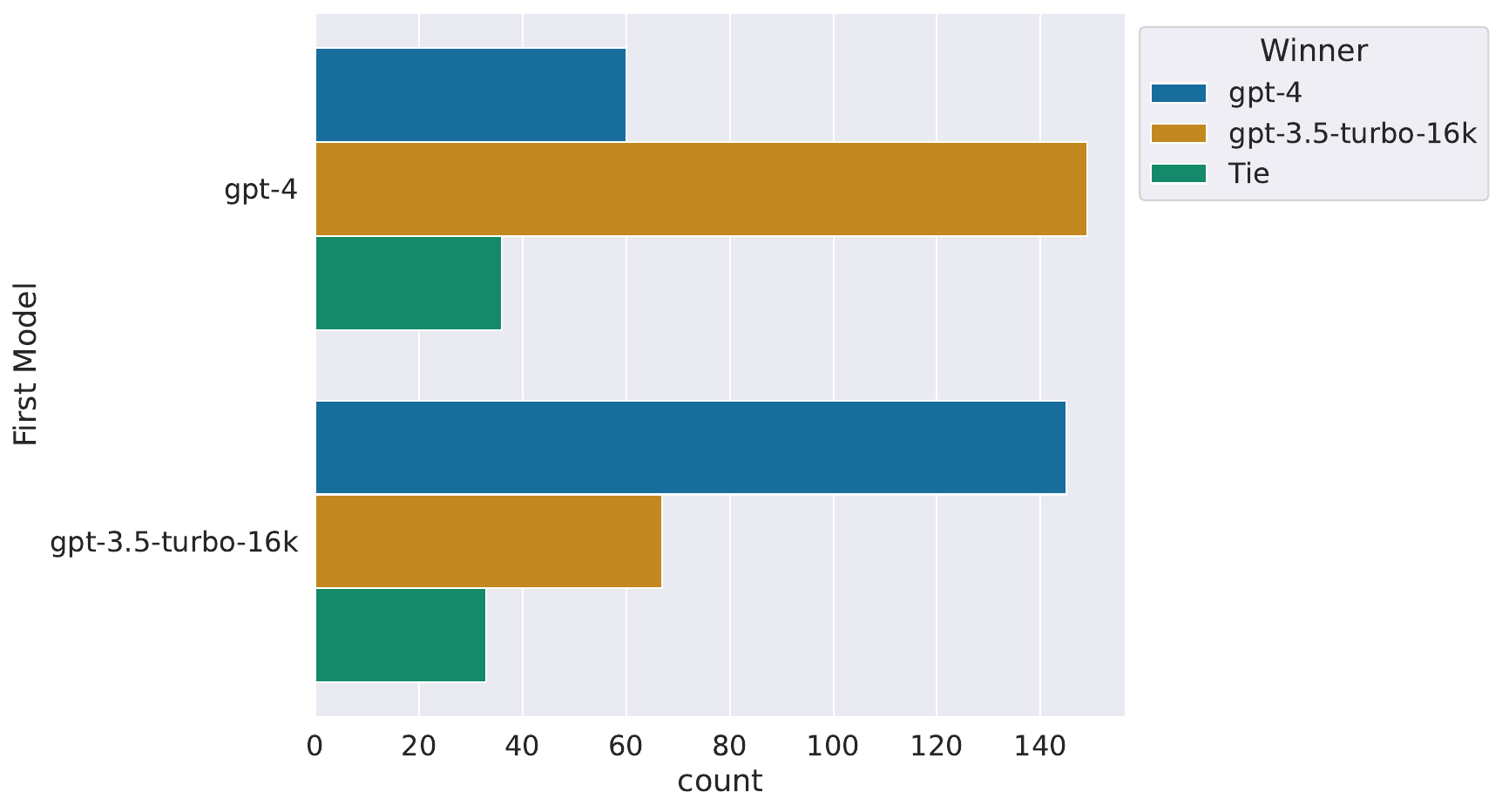}
    \caption{}
    \end{subfigure}
    ~
    \begin{subfigure}[b]{0.48\textwidth}
    \includegraphics[width=.95\linewidth]{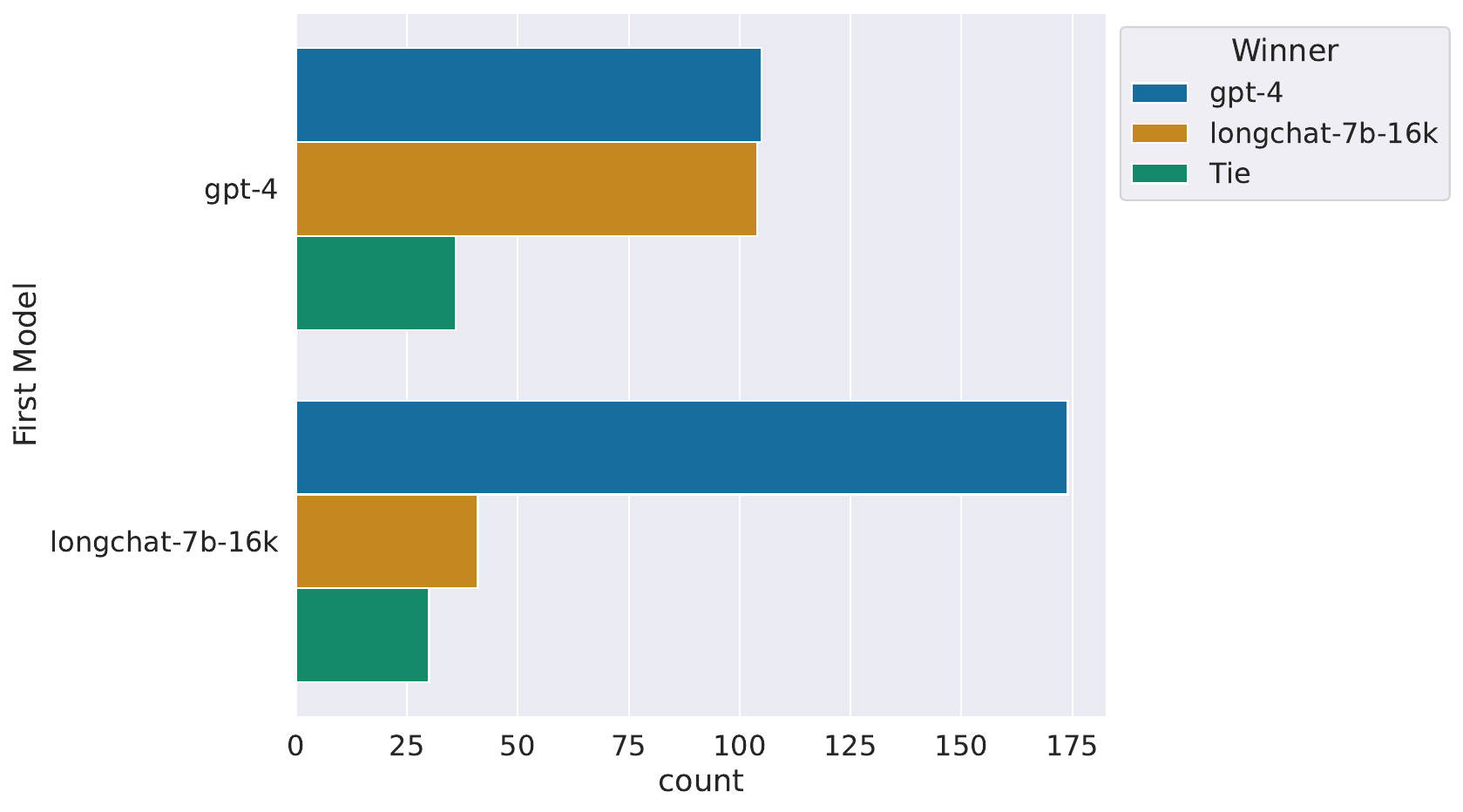}
    \caption{}
    \end{subfigure}

    \begin{subfigure}{0.48\textwidth}

    \includegraphics[width=.95\linewidth]{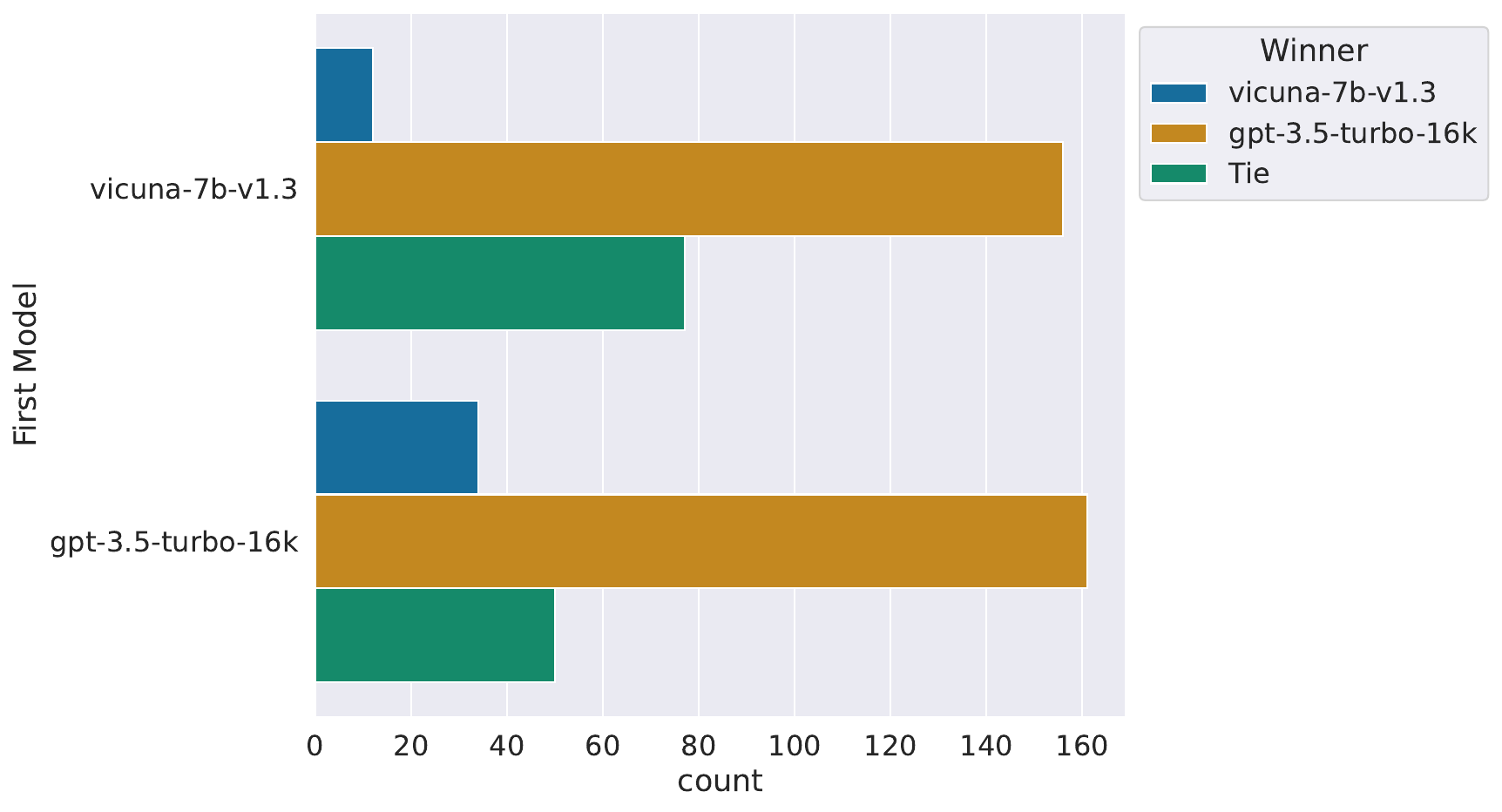}
    \caption{}
    \end{subfigure}
    ~
    \begin{subfigure}[b]{0.48\textwidth}
    \includegraphics[width=.95\linewidth]{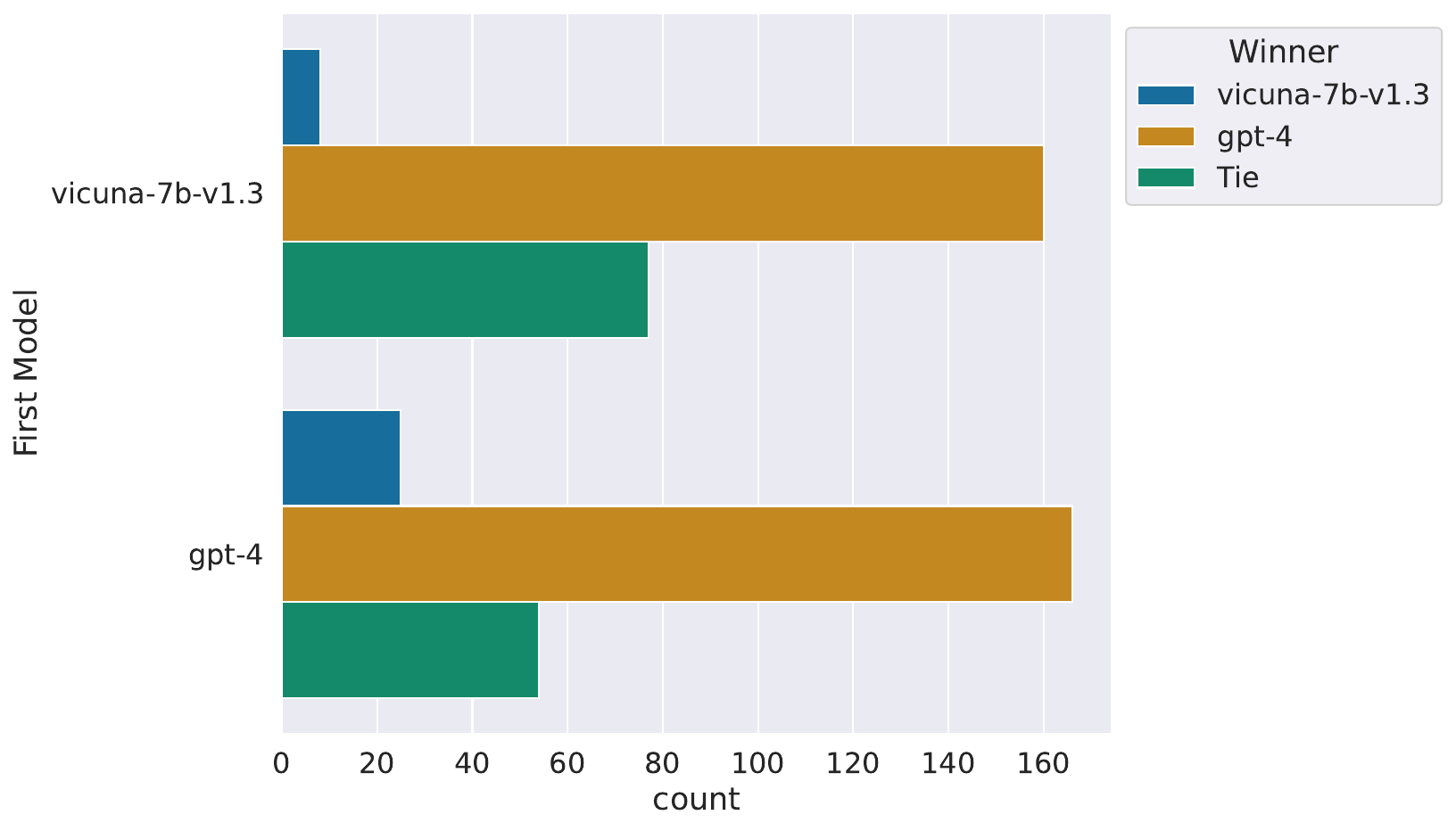}
    \caption{}
    \end{subfigure}

    \begin{subfigure}{0.48\textwidth}

    \includegraphics[width=.95\linewidth]{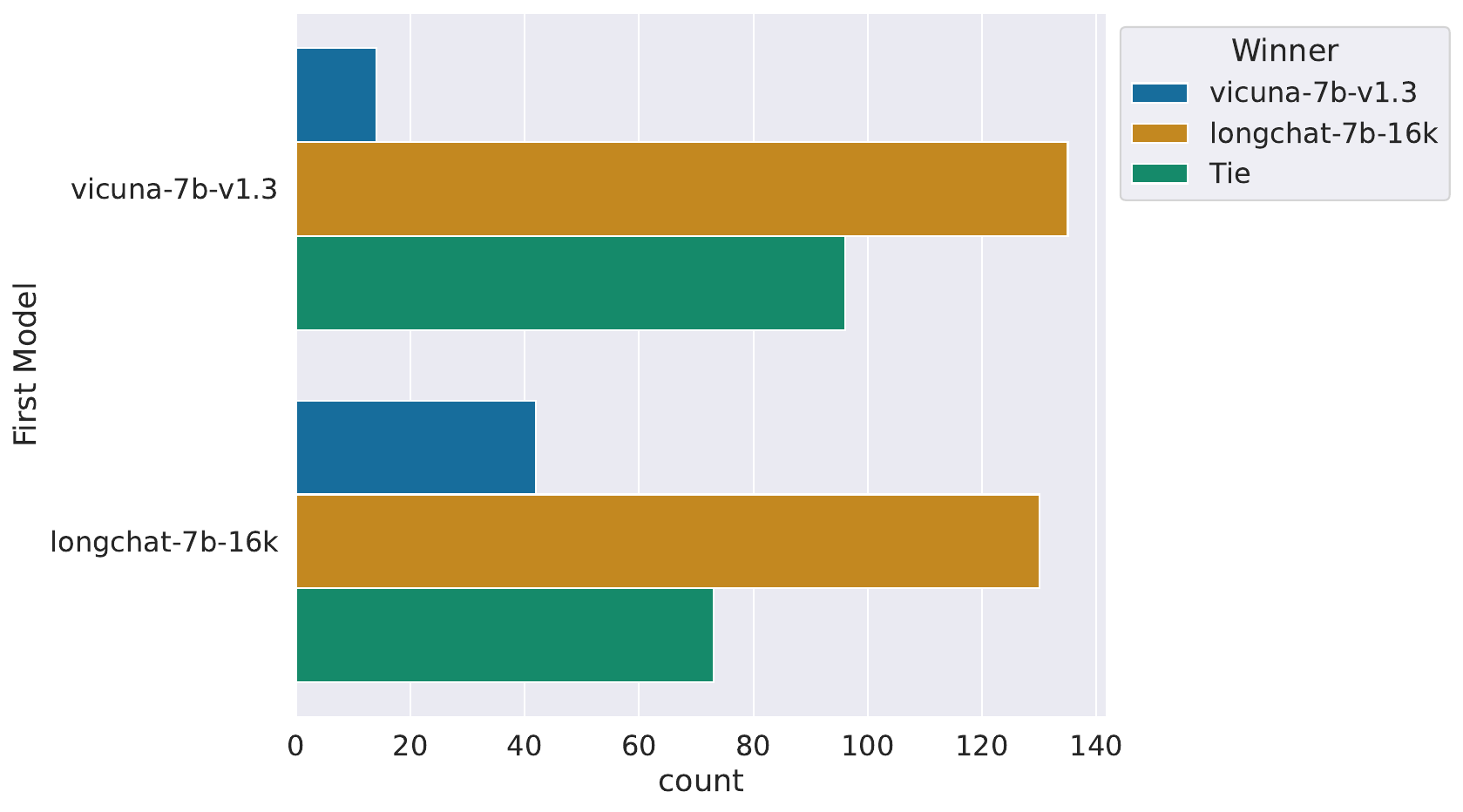}
    \caption{}
    \end{subfigure}
    ~
    \begin{subfigure}[b]{0.48\textwidth}
    \includegraphics[width=.95\linewidth]{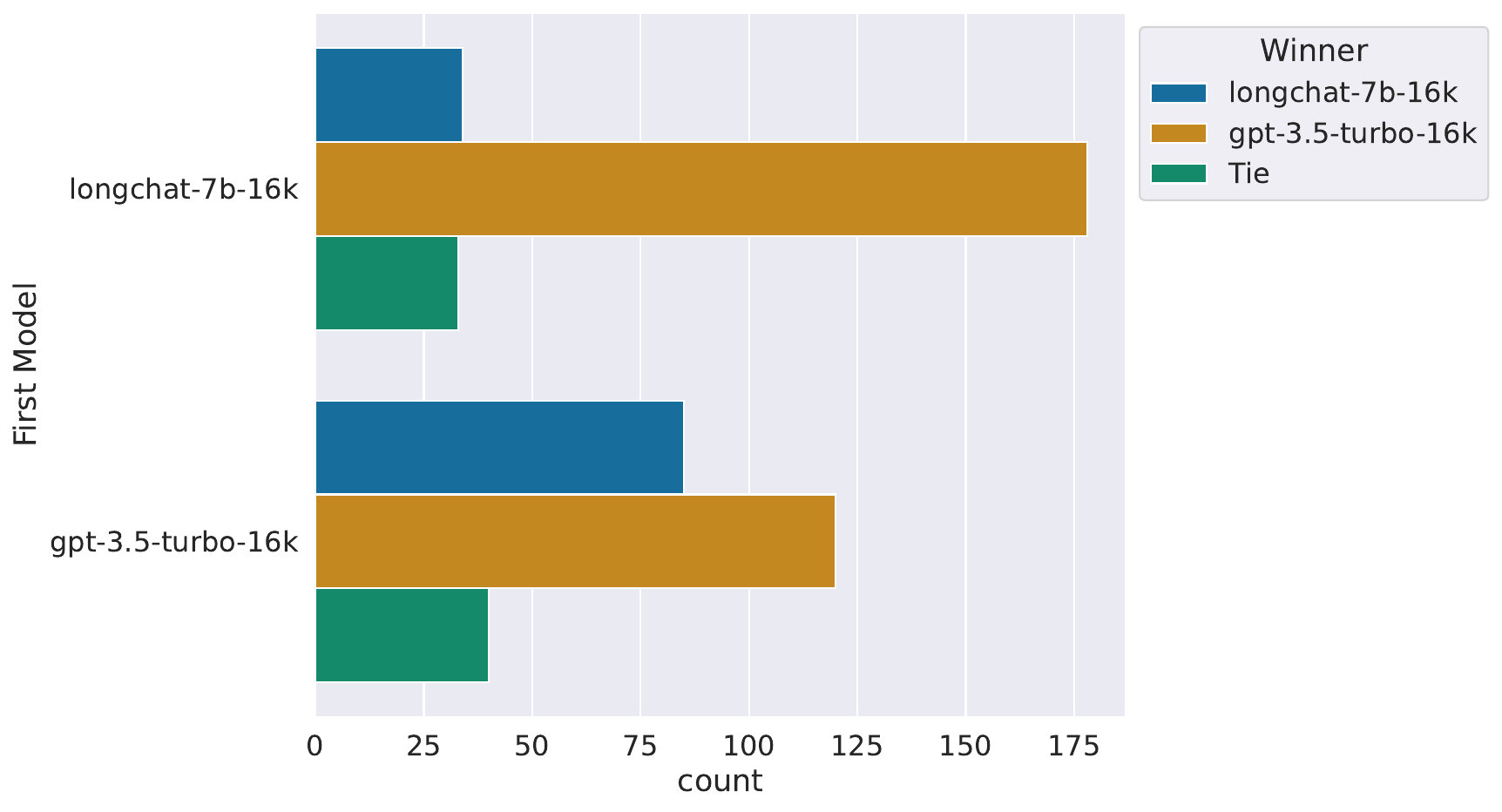}
    \caption{}
    \end{subfigure}

    \caption{Position bias analysis on pairwise comparison protocols for coverage evaluation.\looseness=-1}
    \label{fig:position_bias_pairwise_coverage}
\end{figure*}

\begin{figure*}[t]
    \centering
    \begin{subfigure}{0.48\textwidth}

    \includegraphics[width=.95\linewidth]{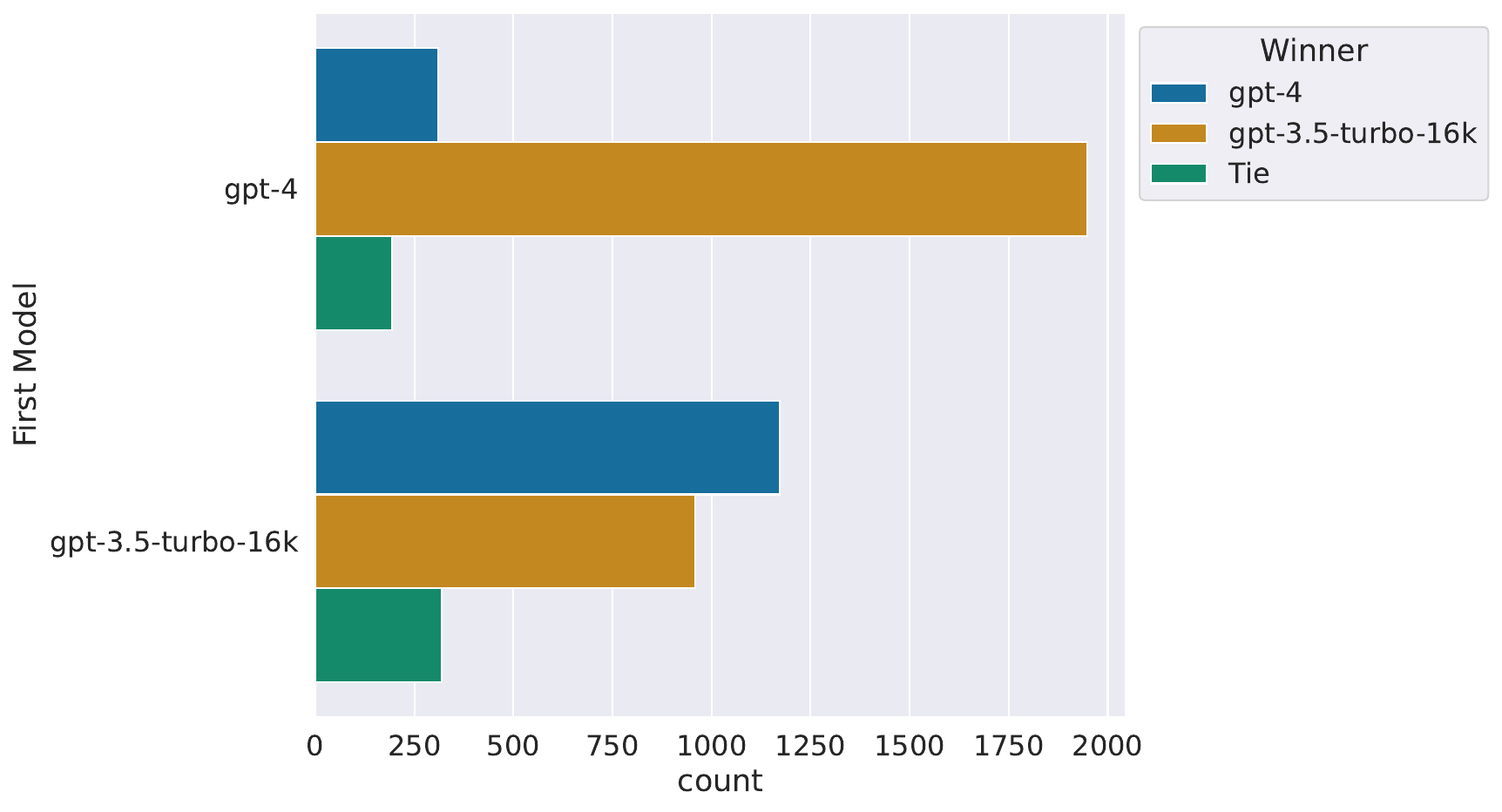}
    \caption{}
    \end{subfigure}
    ~
    \begin{subfigure}[b]{0.48\textwidth}
    \includegraphics[width=.95\linewidth]{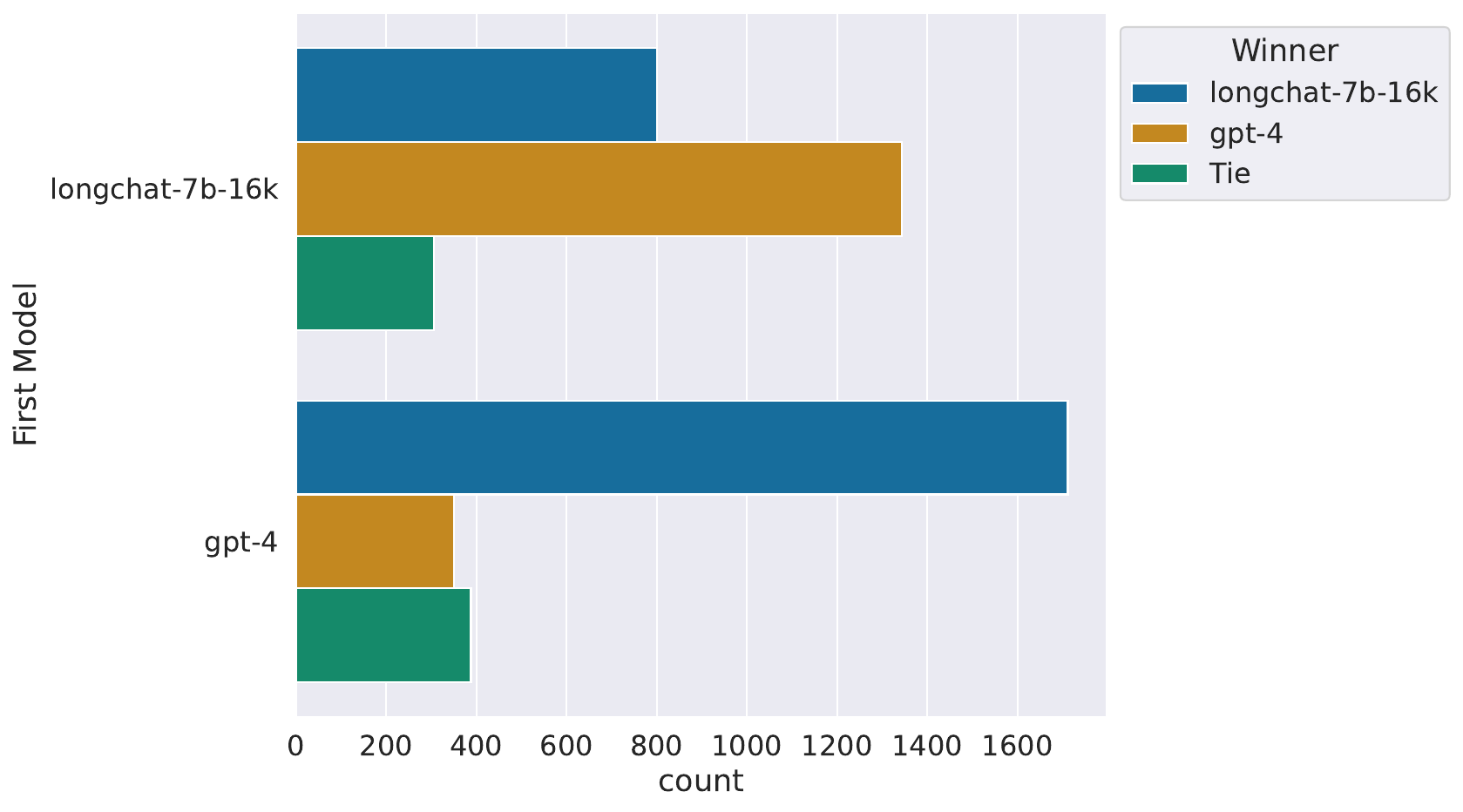}
    \caption{}
    \end{subfigure}

    \begin{subfigure}{0.48\textwidth}

    \includegraphics[width=.95\linewidth]{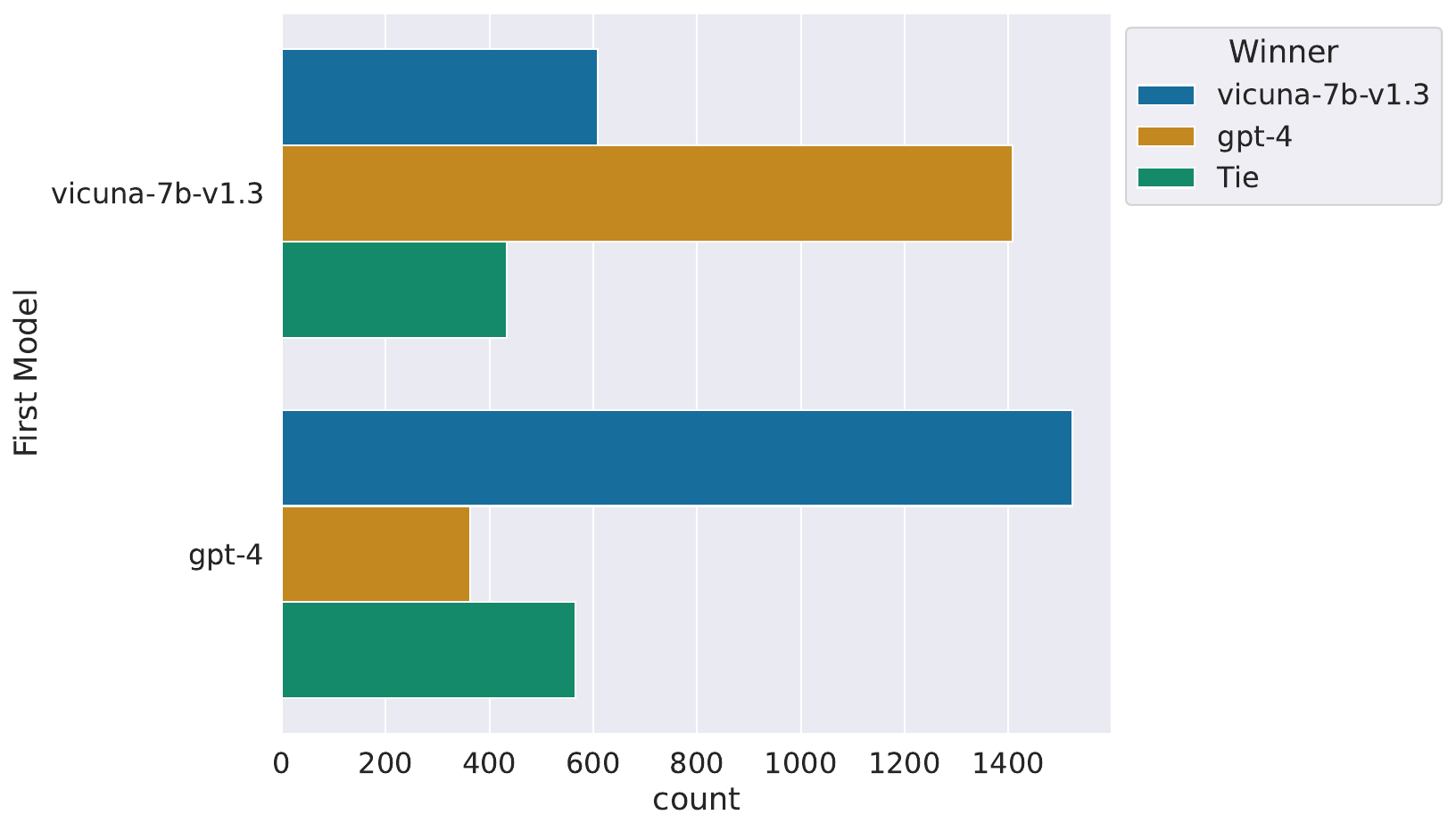}
    \caption{}
    \end{subfigure}
    ~
    \begin{subfigure}[b]{0.48\textwidth}
    \includegraphics[width=.95\linewidth]{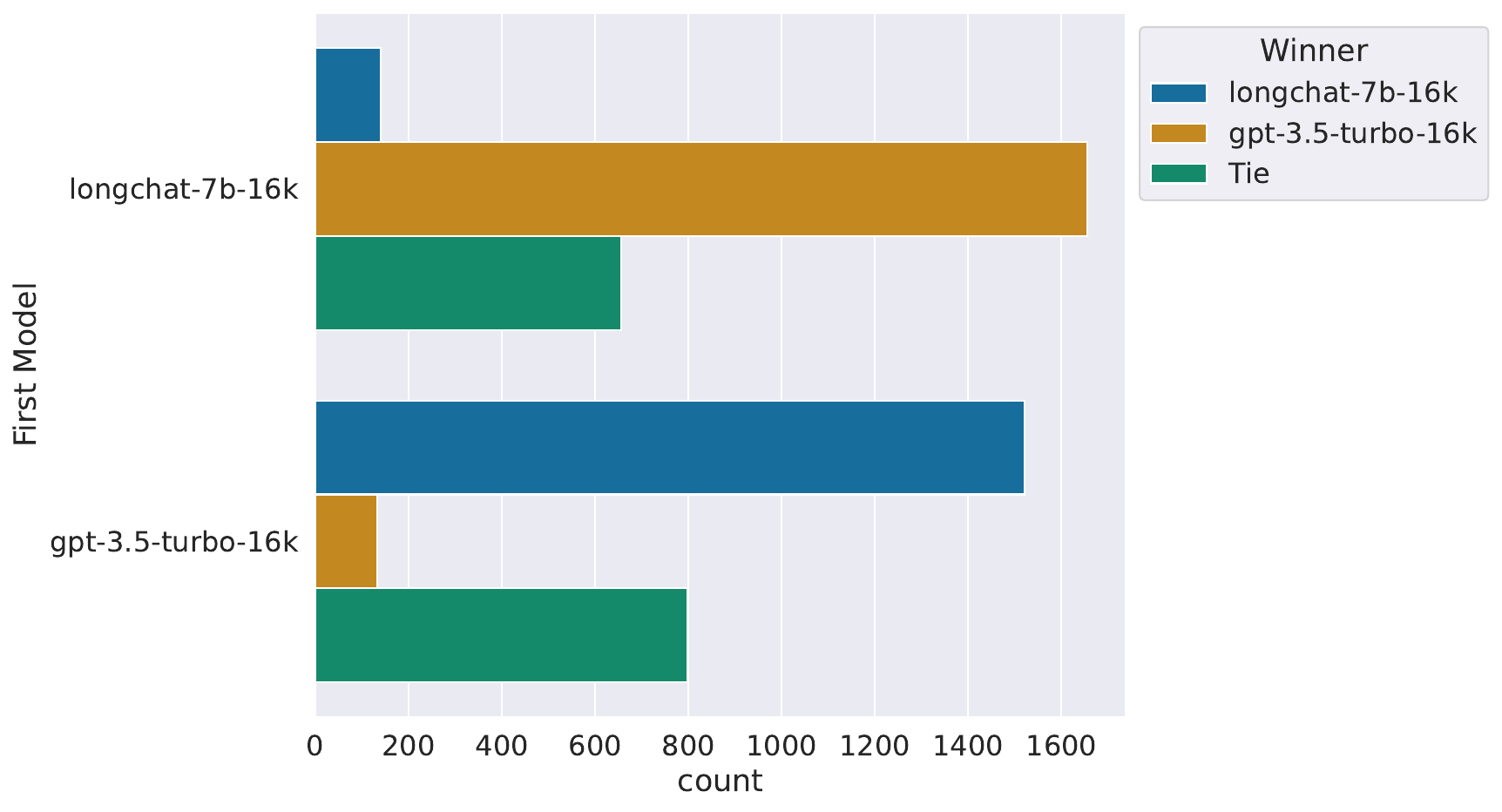}
    \caption{}
    \end{subfigure}

    \begin{subfigure}{0.48\textwidth}

    \includegraphics[width=.95\linewidth]{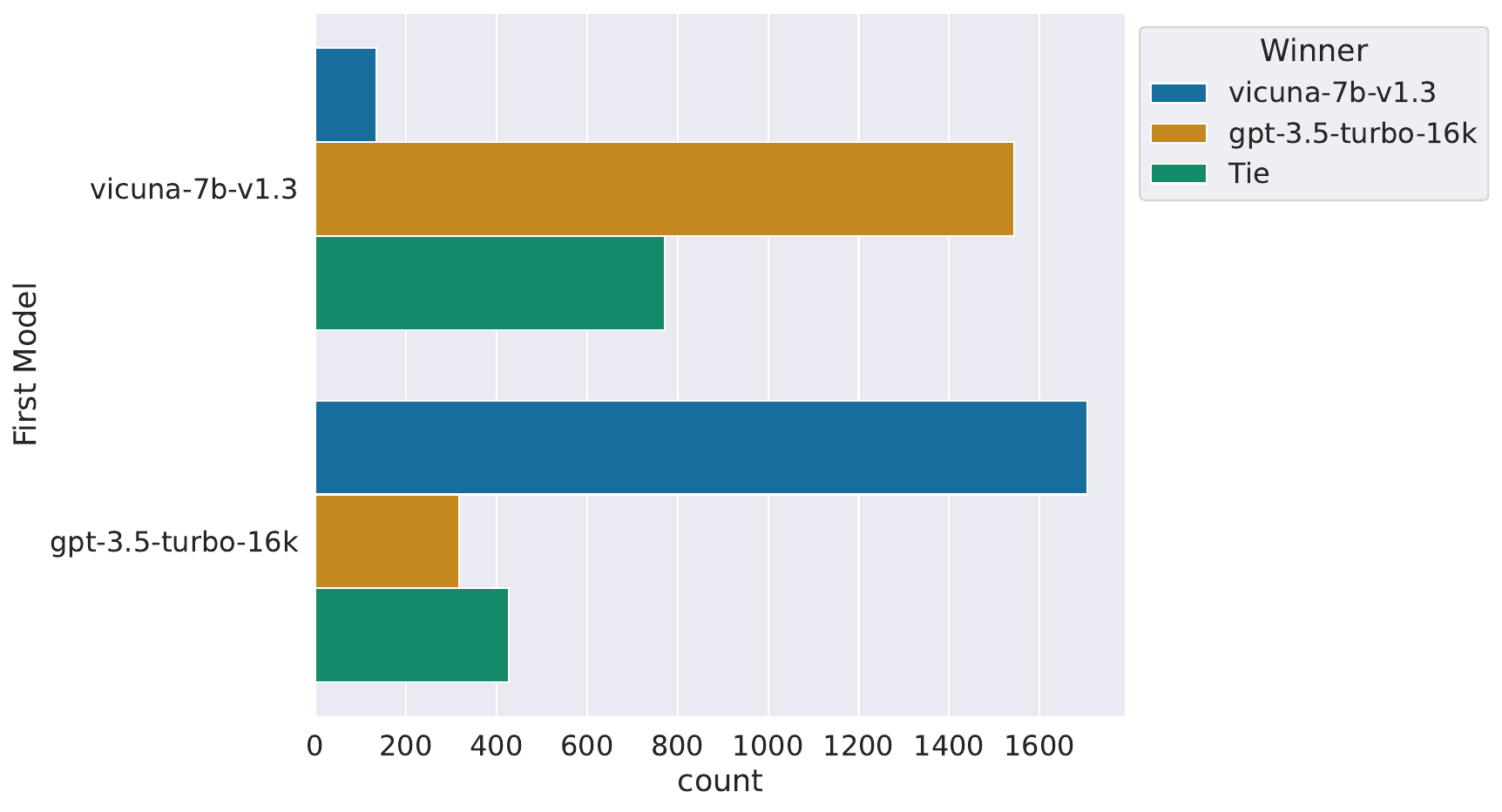}
    \caption{}
    \end{subfigure}
    ~
    \begin{subfigure}[b]{0.48\textwidth}
    \includegraphics[width=.95\linewidth]{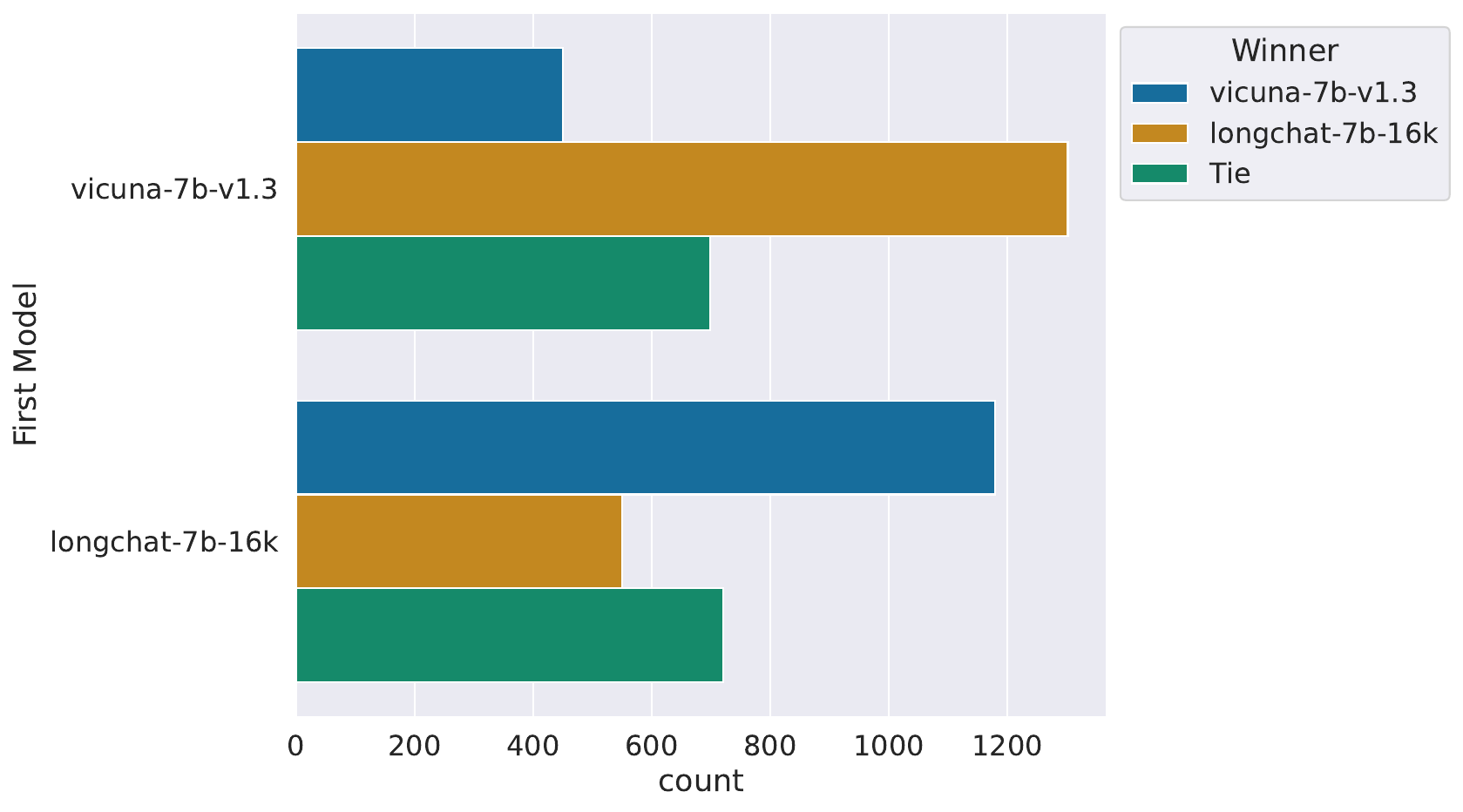}
    \caption{}
    \end{subfigure}

    \caption{Position bias analysis on pairwise comparison protocols for faithfulness evaluation.\looseness=-1}
    \label{fig:position_bias_pairwise_faithfuless}
\end{figure*}

As discussed in \Cref{subsec:req2}, position bias is most relevant to pairwise comparison. \Cref{fig:position_bias_pairwise_coverage} shows the breakdown analysis for coverage evaluation, while the faithfulness evaluation is displayed in \Cref{fig:position_bias_pairwise_faithfuless}. In both coverage and faithfulness evaluation, the evaluator based on \gptfour~  exhibits significant preference towards the second summaries placed in the inputs. In particular, we observe that position bias is most serious when the quality of two summaries is very similar (e.g. (a) in \Cref{fig:position_bias_pairwise_coverage}).

\subsection{Verbosity Bias}
\begin{figure*}[bt]
    \centering
    \begin{subfigure}{0.48\textwidth}

    \includegraphics[width=.95\linewidth]{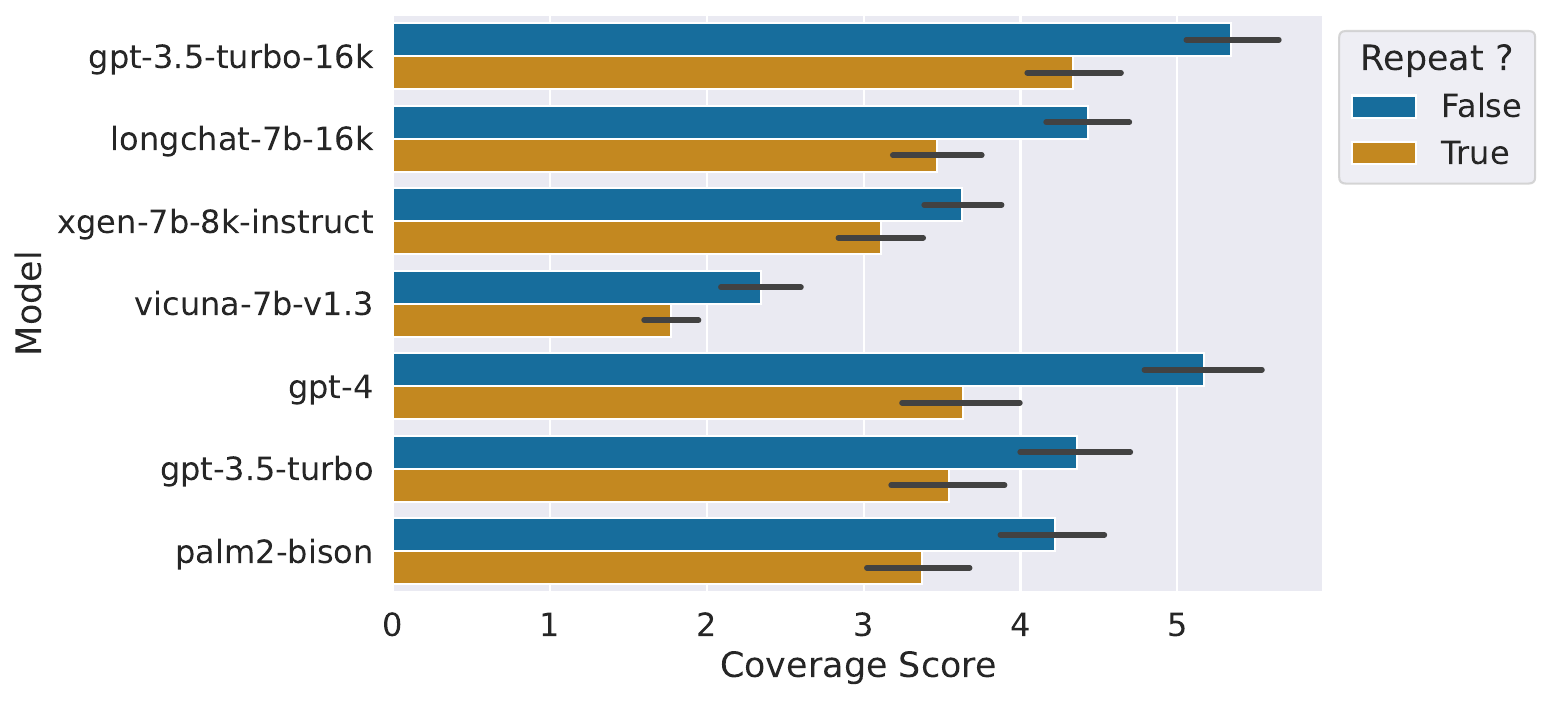}
    
    \end{subfigure}
    ~
    \begin{subfigure}{0.48\textwidth}

    \includegraphics[width=.95\linewidth]{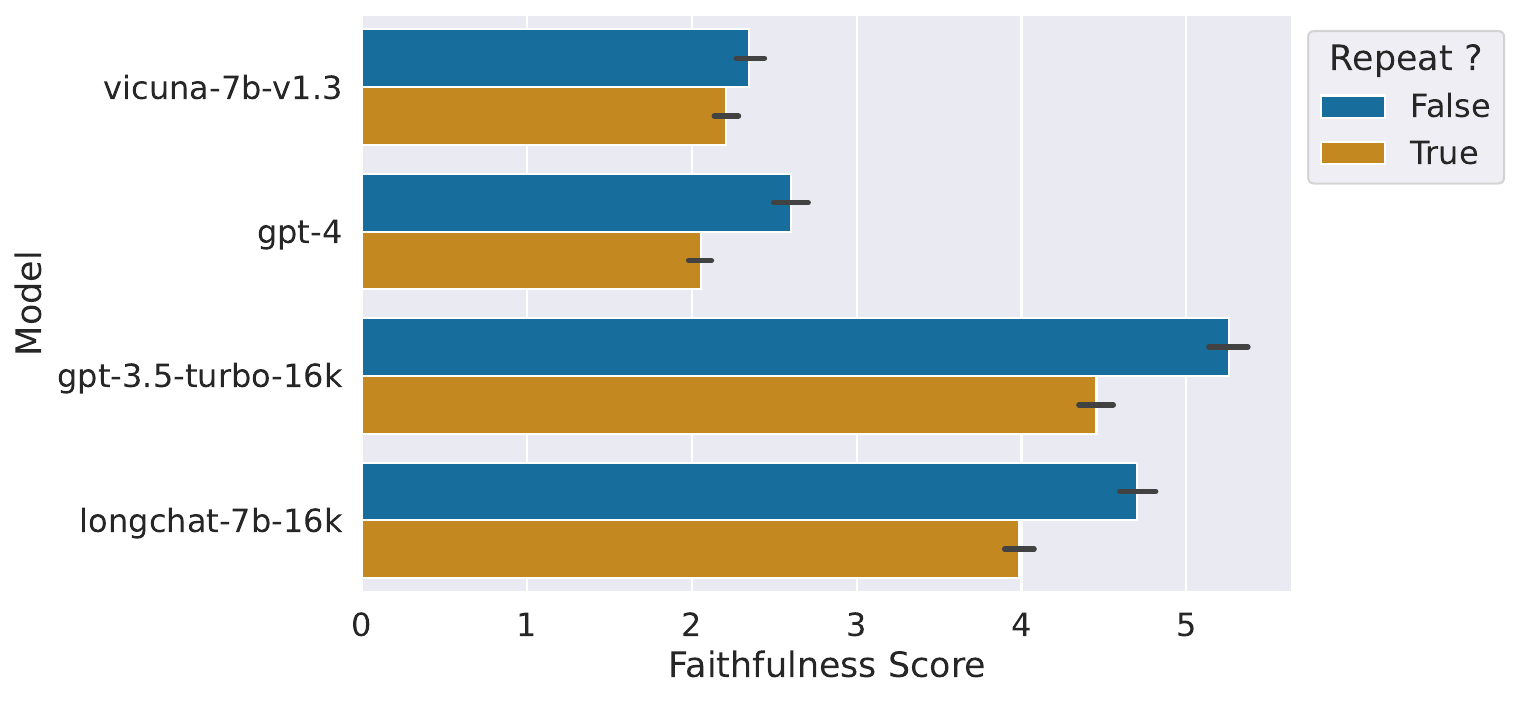}
    
    \end{subfigure}
    \caption{Verbosity analysis using the single-answer grading evaluation protocol. Repeat=False indicates the original summary, while Repeat=True denotes the summary is extended by repeating itself one time.}
    \label{fig:verbosity_sag}
\end{figure*}

As illustrated in \Cref{tab:verbosity_bias_analysis}, pairwise comparison can significantly mitigate the verbosity bias. Hence, in the section, we only show the results for single-answer grading (see \Cref{fig:verbosity_sag}). We see that the GPT-4-based evaluator prefers shorter summaries for all models, no matter when evaluating faithfulness or coverage. The result is surprising since we expect \gptfour~ to prefer longer summaries when performing coverage evaluation.

\section{Topic and Question Distribution}
\label{apx:topic_distribution}
\begin{figure}[bt]
    \centering
    \includegraphics[width=0.55\linewidth]{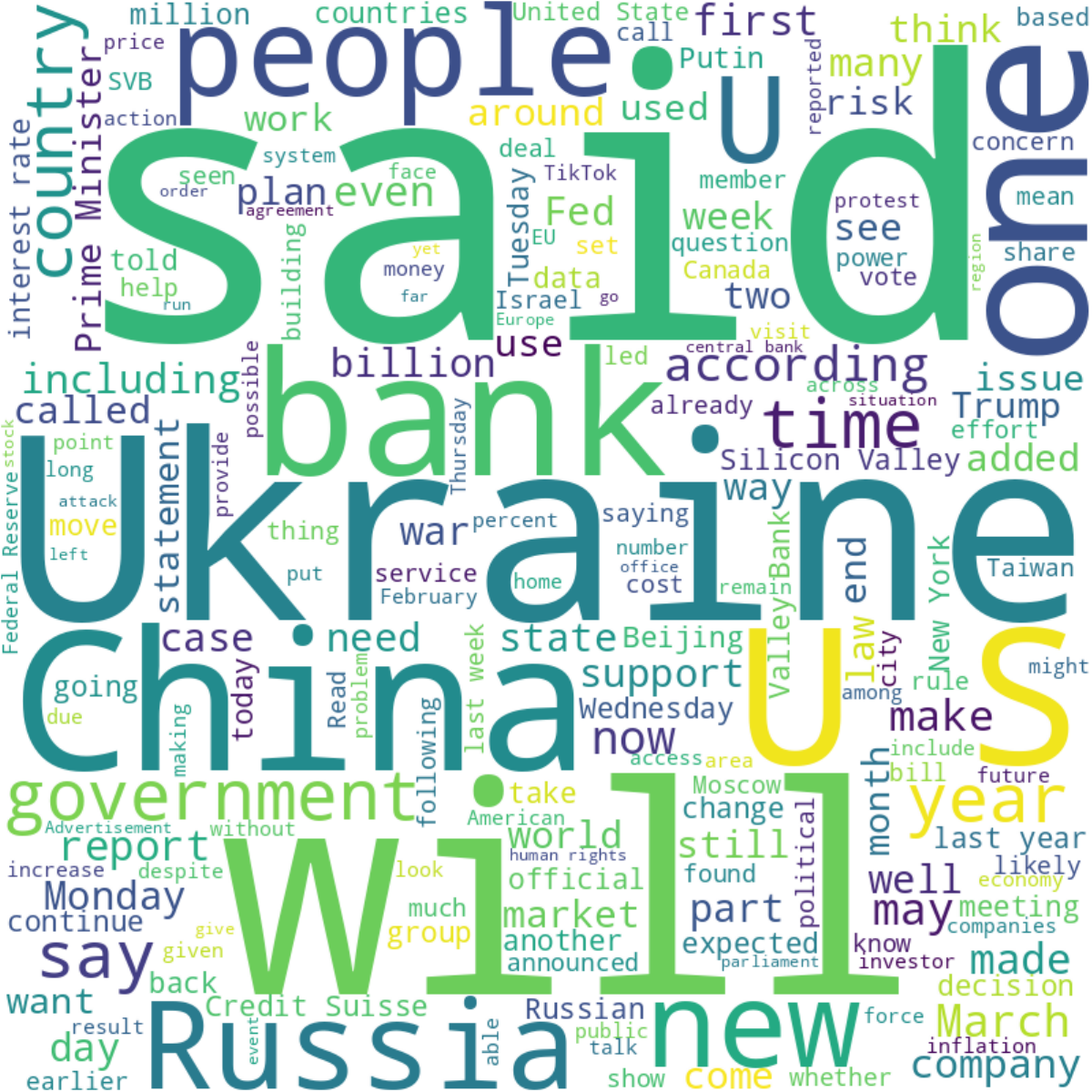}
    \vspace{-2mm}
    \caption{Word cloud representations of the topic distributions over our \datashort~ dataset.} 
    \label{fig:wordcloud}
    \vspace{-5mm}
\end{figure}

\begin{figure}[bt]
    \centering
    \includegraphics[width=0.55\linewidth]{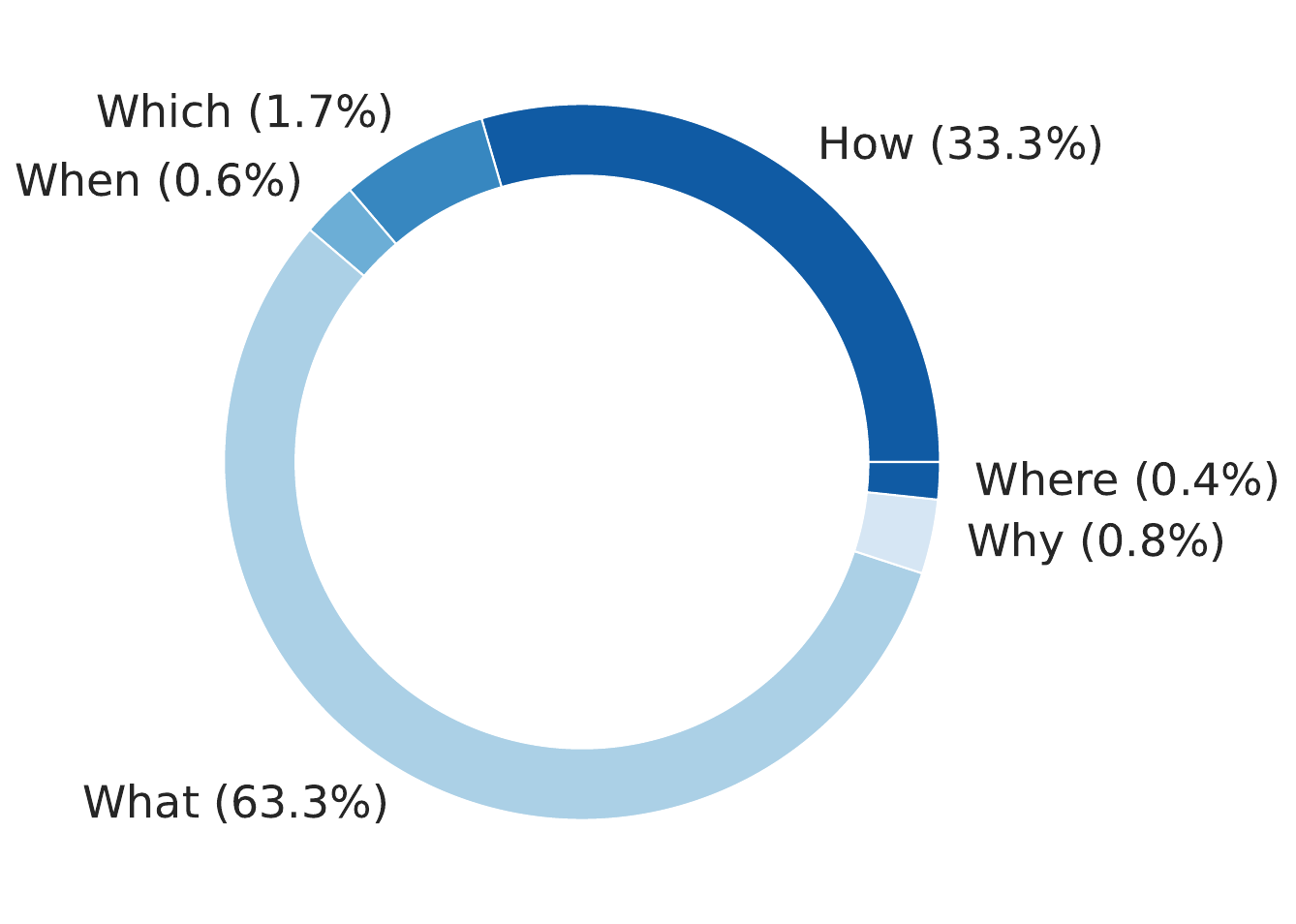}
    \vspace{-2mm}
    \caption{Question distribution of our \datashort~ dataset.} 
    \label{fig:question_distribution}
    \vspace{-5mm}
\end{figure}

\Cref{fig:wordcloud} and \Cref{fig:question_distribution} show the topic distribution and question distribution of our \datashort~ dataset.

